\documentclass[twoside,11pt]{article}
\usepackage{hyperref}       
\usepackage{jair, theapa, rawfonts}

\usepackage[utf8]{inputenc} 
\usepackage[T1]{fontenc}    
\usepackage{url}            
\usepackage{booktabs}       
\usepackage{amsfonts}       
\usepackage{amsmath}
\usepackage{nicefrac}       
\usepackage{microtype}      
\usepackage{xcolor}         
\usepackage{caption}        
\usepackage{subcaption}     
\usepackage{algorithm}
\usepackage{algpseudocode}
\usepackage{tikz}
\usetikzlibrary{shapes.geometric}
\usepackage{multirow}
\usepackage{amsthm}

\usepackage{version}
\excludeversion{kurz}

\newtheorem{definition}{Definition}[section] 
\newtheorem{proposition}[definition]{Proposition} 
\newtheorem{example}[definition]{Example}

\ShortHeadings{What is Fairness?}
 {Bothmann, Peters, \& Bischl}

\begin{document}

\ifdefined\N                                                                
\renewcommand{\N}{\mathds{N}} 
\else \newcommand{\N}{\mathds{N}} \fi 
\newcommand{\Z}{\mathds{Z}} 
\newcommand{\Q}{\mathds{Q}} 
\newcommand{\R}{\mathds{R}} 
\ifdefined\C 
  \renewcommand{\C}{\mathds{C}} 
\else \newcommand{\C}{\mathds{C}} \fi
\newcommand{\continuous}{\mathcal{C}} 
\newcommand{\M}{\mathcal{M}} 
\newcommand{\epsm}{\epsilon_m} 

\newcommand{\setzo}{\{0, 1\}} 
\newcommand{\setmp}{\{-1, +1\}} 
\newcommand{\unitint}{[0, 1]} 

\newcommand{\xt}{\tilde x} 
\newcommand{\argmax}{\operatorname{arg\,max}} 
\newcommand{\argmin}{\operatorname{arg\,min}} 
\newcommand{\argminlim}{\mathop{\mathrm{arg\,min}}\limits} 
\newcommand{\argmaxlim}{\mathop{\mathrm{arg\,max}}\limits} 
\newcommand{\sign}{\operatorname{sign}} 
\newcommand{\I}{\mathbb{I}} 
\newcommand{\order}{\mathcal{O}} 
\newcommand{\pd}[2]{\frac{\partial{#1}}{\partial #2}} 
\newcommand{\floorlr}[1]{\left\lfloor #1 \right\rfloor} 
\newcommand{\ceillr}[1]{\left\lceil #1 \right\rceil} 

\newcommand{\sumin}{\sum\limits_{i=1}^n} 
\newcommand{\sumim}{\sum\limits_{i=1}^m} 
\newcommand{\sumjn}{\sum\limits_{j=1}^n} 
\newcommand{\sumjp}{\sum\limits_{j=1}^p} 
\newcommand{\sumik}{\sum\limits_{i=1}^k} 
\newcommand{\sumkg}{\sum\limits_{k=1}^g} 
\newcommand{\sumjg}{\sum\limits_{j=1}^g} 
\newcommand{\meanin}{\frac{1}{n} \sum\limits_{i=1}^n} 
\newcommand{\meanim}{\frac{1}{m} \sum\limits_{i=1}^m} 
\newcommand{\meankg}{\frac{1}{g} \sum\limits_{k=1}^g} 
\newcommand{\prodin}{\prod\limits_{i=1}^n} 
\newcommand{\prodkg}{\prod\limits_{k=1}^g} 
\newcommand{\prodjp}{\prod\limits_{j=1}^p} 

\newcommand{\one}{\boldsymbol{1}} 
\newcommand{\zero}{\mathbf{0}} 
\newcommand{\id}{\boldsymbol{I}} 
\newcommand{\diag}{\operatorname{diag}} 
\newcommand{\trace}{\operatorname{tr}} 
\newcommand{\spn}{\operatorname{span}} 
\newcommand{\scp}[2]{\left\langle #1, #2 \right\rangle} 
\newcommand{\mat}[1]{\begin{pmatrix} #1 \end{pmatrix}} 
\newcommand{\Amat}{\mathbf{A}} 
\newcommand{\Deltab}{\mathbf{\Delta}} 

\renewcommand{\P}{\mathds{P}} 
\newcommand{\E}{\mathds{E}} 
\newcommand{\var}{\mathsf{Var}} 
\newcommand{\cov}{\mathsf{Cov}} 
\newcommand{\corr}{\mathsf{Corr}} 
\newcommand{\normal}{\mathcal{N}} 
\newcommand{\iid}{\overset{i.i.d}{\sim}} 
\newcommand{\distas}[1]{\overset{#1}{\sim}} 

\newcommand{\Xspace}{\mathcal{X}} 
\newcommand{\Yspace}{\mathcal{Y}} 
\newcommand{\nset}{\{1, \ldots, n\}} 
\newcommand{\pset}{\{1, \ldots, p\}} 
\newcommand{\gset}{\{1, \ldots, g\}} 
\newcommand{\Pxy}{\mathbb{P}_{xy}} 
\newcommand{\Exy}{\mathbb{E}_{xy}} 
\newcommand{\xv}{\mathbf{x}} 
\newcommand{\xtil}{\tilde{\mathbf{x}}} 
\newcommand{\yv}{\mathbf{y}} 
\newcommand{\xy}{(\xv, y)} 
\newcommand{\xvec}{\left(x_1, \ldots, x_p\right)^T} 
\newcommand{\Xmat}{\mathbf{X}} 
\newcommand{\allDatasets}{\mathds{D}} 
\newcommand{\allDatasetsn}{\mathds{D}_n}  
\newcommand{\D}{\mathcal{D}} 
\newcommand{\Dn}{\D_n} 
\newcommand{\Dtrain}{\mathcal{D}_{\text{train}}} 
\newcommand{\Dtest}{\mathcal{D}_{\text{test}}} 
\newcommand{\xyi}[1][i]{\left(\xv^{(#1)}, y^{(#1)}\right)} 
\newcommand{\Dset}{\left( \xyi[1], \ldots, \xyi[n]\right)} 
\newcommand{\defAllDatasetsn}{(\Xspace \times \Yspace)^n} 
\newcommand{\defAllDatasets}{\bigcup_{n \in \N}(\Xspace \times \Yspace)^n} 
\newcommand{\xdat}{\left\{ \xv^{(1)}, \ldots, \xv^{(n)}\right\}} 
\newcommand{\yvec}{\left(y^{(1)}, \hdots, y^{(n)}\right)^T} 
\renewcommand{\xi}[1][i]{\xv^{(#1)}} 
\newcommand{\yi}[1][i]{y^{(#1)}} 
\newcommand{\xivec}{\left(x^{(i)}_1, \ldots, x^{(i)}_p\right)^T} 
\newcommand{\xj}{\xv_j} 
\newcommand{\xjvec}{\left(x^{(1)}_j, \ldots, x^{(n)}_j\right)^T} 
\newcommand{\phiv}{\mathbf{\phi}} 
\newcommand{\phixi}{\mathbf{\phi}^{(i)}} 

\newcommand{\lamv}{\bm{\lambda}} 
\newcommand{\Lam}{\bm{\Lambda}}	 
\newcommand{\preimageInducer}{\left(\defAllDatasets\right)\times\Lam} 
\newcommand{\preimageInducerShort}{\allDatasets\times\Lam} 
\newcommand{\ind}{\mathcal{I}} 

\newcommand{\ftrue}{f_{\text{true}}}  
\newcommand{\ftruex}{\ftrue(\xv)} 
\newcommand{\fx}{f(\xv)} 
\newcommand{\fdomains}{f: \Xspace \rightarrow \R^g} 
\newcommand{\Hspace}{\mathcal{H}} 
\newcommand{\fbayes}{f^{\ast}} 
\newcommand{\fxbayes}{f^{\ast}(\xv)} 
\newcommand{\fkx}[1][k]{f_{#1}(\xv)} 
\newcommand{\fh}{\hat{f}} 
\newcommand{\fxh}{\fh(\xv)} 
\newcommand{\fxt}{f(\xv ~|~ \thetab)} 
\newcommand{\fxi}{f\left(\xv^{(i)}\right)} 
\newcommand{\fxih}{\hat{f}\left(\xv^{(i)}\right)} 
\newcommand{\fxit}{f\left(\xv^{(i)} ~|~ \thetab\right)} 
\newcommand{\fhD}{\fh_{\D}} 
\newcommand{\fhDtrain}{\fh_{\Dtrain}} 
\newcommand{\fhDnlam}{\fh_{\Dn, \lamv}} 
\newcommand{\fhDlam}{\fh_{\D, \lamv}} 
\newcommand{\fhDnlams}{\fh_{\Dn, \lamv^\ast}} 
\newcommand{\fhDlams}{\fh_{\D, \lamv^\ast}} 

\newcommand{\hx}{h(\xv)} 
\newcommand{\hh}{\hat{h}} 
\newcommand{\hxh}{\hat{h}(\xv)} 
\newcommand{\hxt}{h(\xv | \thetab)} 
\newcommand{\hxi}{h\left(\xi\right)} 
\newcommand{\hxit}{h\left(\xi ~|~ \thetab\right)} 
\newcommand{\hbayes}{h^{\ast}} 
\newcommand{\hxbayes}{h^{\ast}(\xv)} 

\newcommand{\yh}{\hat{y}} 
\newcommand{\yih}{\hat{y}^{(i)}} 
\newcommand{\resi}{\yi- \yih}

\newcommand{\thetah}{\hat{\theta}} 
\newcommand{\thetab}{\bm{\theta}} 
\newcommand{\thetabh}{\bm{\hat\theta}} 
\newcommand{\thetat}[1][t]{\thetab^{[#1]}} 
\newcommand{\thetatn}[1][t]{\thetab^{[#1 +1]}} 
\newcommand{\thetahDnlam}{\thetabh_{\Dn, \lamv}} 
\newcommand{\thetahDlam}{\thetabh_{\D, \lamv}} 
\newcommand{\mint}{\min_{\thetab \in \Theta}} 
\newcommand{\argmint}{\argmin_{\thetab \in \Theta}} 

\newcommand{\pdf}{p} 
\newcommand{\pdfx}{p(\xv)} 
\newcommand{\pixt}{\pi(\xv~|~ \thetab)} 
\newcommand{\pixit}{\pi\left(\xi ~|~ \thetab\right)} 
\newcommand{\pixii}{\pi(\xi)} 
\newcommand{\pii}{\pi^{(i)}} 

\newcommand{\pdfxy}{p(\xv,y)} 
\newcommand{\pdfxyt}{p(\xv, y ~|~ \thetab)} 
\newcommand{\pdfxyit}{p\left(\xi, \yi ~|~ \thetab\right)} 

\newcommand{\pdfxyk}[1][k]{p(\xv | y= #1)} 
\newcommand{\lpdfxyk}[1][k]{\log p(\xv | y= #1)} 
\newcommand{\pdfxiyk}[1][k]{p\left(\xi | y= #1 \right)} 

\newcommand{\pik}[1][k]{\pi_{#1}} 
\newcommand{\lpik}[1][k]{\log \pi_{#1}} 
\newcommand{\pit}{\pi(\thetab)} 

\newcommand{\post}{\P(y = 1 ~|~ \xv)} 
\newcommand{\postk}[1][k]{\P(y = #1 ~|~ \xv)} 
\newcommand{\pidomains}{\pi: \Xspace \rightarrow \unitint} 
\newcommand{\pibayes}{\pi^{\ast}} 
\newcommand{\pixbayes}{\pi^{\ast}(\xv)} 
\newcommand{\pix}{\pi(\xv)} 
\newcommand{\pikx}[1][k]{\pi_{#1}(\xv)} 
\newcommand{\pikxt}[1][k]{\pi_{#1}(\xv ~|~ \thetab)} 
\newcommand{\pixh}{\hat \pi(\xv)} 
\newcommand{\pikxh}[1][k]{\hat \pi_{#1}(\xv)} 
\newcommand{\pixih}{\hat \pi(\xi)} 
\newcommand{\pikxih}[1][k]{\hat \pi_{#1}(\xi)} 
\newcommand{\pdfygxt}{p(y ~|~\xv, \thetab)} 
\newcommand{\pdfyigxit}{p\left(\yi ~|~\xi, \thetab\right)} 
\newcommand{\lpdfygxt}{\log \pdfygxt } 
\newcommand{\lpdfyigxit}{\log \pdfyigxit} 

\newcommand{\bayesrulek}[1][k]{\frac{\P(\xv | y= #1) \P(y= #1)}{\P(\xv)}} 
\newcommand{\muk}{\bm{\mu_k}} 

\newcommand{\eps}{\epsilon} 
\newcommand{\epsi}{\epsilon^{(i)}} 
\newcommand{\epsh}{\hat{\epsilon}} 
\newcommand{\yf}{y \fx} 
\newcommand{\yfi}{\yi \fxi} 
\newcommand{\Sigmah}{\hat \Sigma} 
\newcommand{\Sigmahj}{\hat \Sigma_j} 

\newcommand{\Lyf}{L\left(y, f\right)} 
\newcommand{\Lxy}{L\left(y, \fx\right)} 
\newcommand{\Lxyi}{L\left(\yi, \fxi\right)} 
\newcommand{\Lxyt}{L\left(y, \fxt\right)} 
\newcommand{\Lxyit}{L\left(\yi, \fxit\right)} 
\newcommand{\Lxym}{L\left(\yi, f\left(\bm{\tilde{x}}^{(i)} ~|~ \thetab\right)\right)} 
\newcommand{\Lpixy}{L\left(y, \pix\right)} 
\newcommand{\Lpixyi}{L\left(\yi, \pixii\right)} 
\newcommand{\Lpixyt}{L\left(y, \pixt\right)} 
\newcommand{\Lpixyit}{L\left(\yi, \pixit\right)} 
\newcommand{\Lhxy}{L\left(y, \hx\right)} 
\newcommand{\Lr}{L\left(r\right)} 
\newcommand{\lone}{|y - \fx|} 
\newcommand{\ltwo}{\left(y - \fx\right)^2} 
\newcommand{\lbernoullimp}{\ln(1 + \exp(-y \cdot \fx))} 
\newcommand{\lbernoullizo}{- y \cdot \fx + \log(1 + \exp(\fx))} 
\newcommand{\lcrossent}{- y \log \left(\pix\right) - (1 - y) \log \left(1 - \pix\right)} 
\newcommand{\lbrier}{\left(\pix - y \right)^2} 
\newcommand{\risk}{\mathcal{R}} 
\newcommand{\riskbayes}{\mathcal{R}^\ast}
\newcommand{\riskf}{\risk(f)} 
\newcommand{\riskdef}{\E_{y|\xv}\left(\Lxy \right)} 
\newcommand{\riskt}{\mathcal{R}(\thetab)} 
\newcommand{\riske}{\mathcal{R}_{\text{emp}}} 
\newcommand{\riskeb}{\bar{\mathcal{R}}_{\text{emp}}} 
\newcommand{\riskef}{\riske(f)} 
\newcommand{\risket}{\mathcal{R}_{\text{emp}}(\thetab)} 
\newcommand{\riskr}{\mathcal{R}_{\text{reg}}} 
\newcommand{\riskrt}{\mathcal{R}_{\text{reg}}(\thetab)} 
\newcommand{\riskrf}{\riskr(f)} 
\newcommand{\riskrth}{\hat{\mathcal{R}}_{\text{reg}}(\thetab)} 
\newcommand{\risketh}{\hat{\mathcal{R}}_{\text{emp}}(\thetab)} 
\newcommand{\LL}{\mathcal{L}} 
\newcommand{\LLt}{\mathcal{L}(\thetab)} 
\newcommand{\LLtx}{\mathcal{L}(\thetab | \xv)} 
\newcommand{\logl}{\ell} 
\newcommand{\loglt}{\logl(\thetab)} 
\newcommand{\logltx}{\logl(\thetab | \xv)} 
\newcommand{\errtrain}{\text{err}_{\text{train}}} 
\newcommand{\errtest}{\text{err}_{\text{test}}} 
\newcommand{\errexp}{\overline{\text{err}_{\text{test}}}} 

\newcommand{\thx}{\thetab^T \xv} 
\newcommand{\olsest}{(\Xmat^T \Xmat)^{-1} \Xmat^T \yv} 

\newcommand{\xti}{\xtil^{(i)}}
\newcommand{\yti}{\tilde{y}^{(i)}}
\newcommand{\yt}{\tilde{y}}
\newcommand{\xtilde}{\tilde{\xv}}

\newcommand{\pitil}{\varphi}
\newcommand{\pitilh}{\hat{\pitil}}
\newcommand{\piti}{\pitil^{(i)}}
\newcommand{\pitj}{\pitil^{(j)}}
\newcommand{\pitifh}{\hat{\pitil}(\cdot)}
\newcommand{\pitxtih}{\hat{\pitil}(\xti)} 
\newcommand{\pitxih}{\hat{\pitil}(\xi)} 
\newcommand{\pitxti}{\pitil(\xti)}
\newcommand{\pitxi}{\pitil(\xi)}
\newcommand{\pixv}{\pi(\xv)}

\newcommand{\spiti}{s(\piti)}
\newcommand{\spitxtih}{s(\pitxtih)}
\newcommand{\spitxih}{s(\pitxih)}
\newcommand{\spitxti}{s(\pitxti)}
\newcommand{\spitxi}{s(\pitxi)}

\newcommand{\spii}{s(\pii)}
\newcommand{\spixih}{s(\pixih)}
\newcommand{\spixii}{s(\pixii)}

\newcommand{\psih}{\hat{\psi}}
\newcommand{\psihi}{\psih^{(i)}}
\newcommand{\psii}{\psi^{(i)}}
\newcommand{\psij}{\psi^{(j)}}

\newcommand{\psif}{\psi(\cdot)}

\newcommand{\spsii}{s(\psii)}
\newcommand{\spsixii}{s(\psi(\xfindi))}
\newcommand{\spsixih}{s(\psih(\xfindi))}

\newcommand{\ti}{t^{(i)}}
\newcommand{\tj}{t^{(j)}}
\newcommand{\wi}{m^{(i)}}
\newcommand{\w}{m}
\newcommand{\W}{M}
\newcommand{\wj}{m^{(j)}}
\newcommand{\zi}{z^{(i)}}
\newcommand{\vi}{v^{(i)}}
\newcommand{\mui}{\mu^{(i)}}

\newcommand{\sfu}{s(\cdot)}
\newcommand{\pij}{\pi^{(j)}}
\newcommand{\pif}{\pi(\cdot)}
\newcommand{\pih}{\hat{\pi}}
\newcommand{\pihi}{\pih^{(i)}}
\newcommand{\pihj}{\pih^{(j)}}
\newcommand{\pihf}{\pih(\cdot)}
\newcommand{\pitf}{\pitil(\cdot)}
\newcommand{\pith}{\hat{\pitil}}
\newcommand{\pithf}{\hat{\pitil}(\cdot)}
\newcommand{\pithi}{\hat{\pitil}^{(i)}}
\newcommand{\pithj}{\hat{\pitil}^{(j)}}

\newcommand{\xyti}[1][i]{\left(\xtil^{(#1)}, \yt^{(#1)}\right)} 
\newcommand{\Dsett}{\left( \xyti[1], \ldots, \xyti[n]\right)} 
\newcommand{\Dt}{\tilde{\mathcal{D}}} 

\newcommand{\Xspacet}{\tilde{\Xspace}}

\newcommand{\xfindi}{\xv_F^{(i)}}
\newcommand{\xfindj}{\xv_F^{(j)}}
\newcommand{\yfindi}{y_F^{(i)}}
\newcommand{\yfind}{y_F}

\newcommand{\Cv}{\boldsymbol{C}}
\newcommand{\cvi}{\boldsymbol{c}^{(i)}}
\newcommand{\ri}{r^{(i)}}
\newcommand{\xpi}{x_P^{(i)}}
\newcommand{\xhpi}{\tilde{x}_P^{(i)}}
\newcommand{\xhdi}{\tilde{x}_D^{(i)}}
\newcommand{\qwi}{q_w^{(i)}}

\newcommand{\lb}[1]{\textcolor{blue}{[L: #1]}}
\newcommand{\kp}[1]{\textcolor{blue}{[K: #1]}}
\newcommand{\bp}[1]{\textcolor{blue}{$\rightarrow$ #1 \\}}

\newcommand{\myemph}[1]{\textit{#1}}

\newcommand{\mydef}[1]{\paragraph{#1}}

\title{What Is Fairness? On the Role of Protected Attributes and Fictitious Worlds}

 \author{\name Ludwig Bothmann \email ludwig.bothmann@lmu.de \\
        \addr LMU Munich, Germany\\
        Munich Center for Machine Learning (MCML) \\
        \AND
        \name Kristina Peters \email kristina.peters@jura.uni-muenchen.de \\
        \addr LMU Munich, Germany\\
        \AND
        \name Bernd Bischl \email bernd.bischl@stat.uni-muenchen.de \\
        \addr LMU Munich, Germany\\
        Munich Center for Machine Learning (MCML) 
        }


\maketitle

\begin{abstract}
A growing body of literature in fairness-aware machine learning (fairML) aims to mitigate machine learning (ML)-related unfairness in automated decision-making (ADM) by defining metrics that measure fairness of an ML model and by proposing methods to ensure that trained ML models achieve low scores on these metrics.
However, the underlying concept of fairness, i.e., the question of what fairness is, is rarely discussed, leaving a significant gap between centuries of philosophical discussion and the recent adoption of the concept in the ML community.
In this work, we try to bridge this gap by formalizing a consistent concept of fairness and by translating the philosophical considerations into a formal framework for the training and evaluation of ML models in ADM systems.
We argue that fairness problems can arise even without the presence of protected attributes (PAs), and point out that fairness and predictive performance are not irreconcilable opposites, but that the latter is necessary to achieve the former.
Furthermore, we argue why and how causal considerations are necessary when assessing fairness in the presence of PAs by proposing a fictitious, normatively desired (FiND) world in which PAs have no causal effects.
In practice, this FiND world must be approximated by a warped world in which the causal effects of the PAs are removed from the real-world data.
Finally, we achieve greater linguistic clarity in the discussion of fairML. We outline algorithms for practical applications and present illustrative experiments on COMPAS data.
\end{abstract}

\section{Introduction}
\label{sec:intro}

The machine learning (ML) community has produced numerous contributions on the topic of fairness-aware ML (fairML) in recent years.
However, a fundamental question remains: \textit{What is fairness?} This question is not easily answered and is often circumvented; instead of asking ``what is fairness'', the questions of ``how to measure fairness of ML models'' and ``how to make ML models fair'' are pursued. 

This paper does not intend to criticize individual approaches that address those latter questions and, in doing so, often propose important solutions for specific problems.
Rather, the aim is to make explicit the premises that 
underlie the various understandings of fairness and the approaches to solving fairness problems. In doing so, a broadly consistent understanding can be based on a rich foundation in the history of philosophy. Subsequently, we show that the conception of fairness depends on multilayered normative evaluations;
any discussion of fairML relies on adopting those normative stipulations.
The basis for fair decisions is always the question of the \textit{equality} of the people treated \textit{with respect to the subject matter} concerned. 
On this basis, a decision rule is to be established, which in turn can be adapted to the concrete needs as a result of normative stipulations.
Based on this essential concept of fairness, we turn to the questions of to what extent ML models can induce unfair treatments in automated decision-making (ADM), and of how to implement these normative stipulations both in training an ML model and in using its predictions in ADM. 

\subsection{Our Contributions}

    In this paper, we formalize a \myemph{consistent concept of fairness} derived from philosophical considerations and translate it into a formal framework for the design of ML models in ADM systems, hence bridging the gap between centuries of philosophical discussion and the recent adoption in the ML community (Section \ref{sec:defs}). 
  We precisely delineate fairML's contribution to questions of fairness in ADM from the responsibilities of other scientific fields as well as from the socio-political discussion and from legislative decisions; this distinction offers \myemph{greater linguistic clarity} for the discussion of algorithmic fairness (Sections \ref{sec:defs} and \ref{sec:adm}).
  We argue that an ML model cannot be unfair per se because a predictive model alone does not make decisions or execute material actions. Rather, 
  \myemph{fairness problems can arise} using ML models in ADM systems if the model is not individually well-calibrated -- \myemph{even if no protected attributes (PAs) are concerned}. Hence, we point out that predictive performance is paramount for fairness rather than a tradeoff to it (Section \ref{sec:ml_ohne_pa}).
  \myemph{In the presence of PAs}, we tackle the goal of PA-neutrality (leaving PA-focus to future research, see Section \ref{sec:equality} for the distinction of these two concepts): a fictitious, normatively desired (FiND) world is conceived, where the PAs have no causal effects. This FiND world must be approximated by a warped world, and ML models are to be trained and evaluated in the warped world. We emphasize that \myemph{fairness criteria must make causal considerations} (Section \ref{sec:ml_mit_pa}).
In addition, our concept offers an explanation for the possibility for fair but unequal treatment of unequal individuals -- also called vertical equity \shortcite{black_algorithmic_2022}. 
Finally, we outline first \myemph{algorithms for practical applications} and show illustrative \myemph{empirical results} on COMPAS data (Section \ref{sec:implementation}).
We focus on the theoretical problems and fundamental challenges surrounding fairML. While we investigate and diagnose these thoroughly, we also indicate how these problems might be overcome. We do not present final solutions to all open questions but put the necessary emphasis on clarifying the structural problems in fairML to enable future research to investigate proper solutions that could help mitigate ethical issues of ADM. 
    

\subsection{Motivating Examples}

Our concept aligns well with the notions of \myemph{substantive equality} and \myemph{bias transforming fairness metrics} as described, e.g., by \shortciteA{wachter_bias_2021}; in Section \ref{sec:comp_other_concepts} we compare our framework with these notions. Before presenting our framework in detail, we illustrate our understanding with two examples:

\paragraph{Recidivism Prediction} Consider a Person of Color (PoC) who is charged with a crime and whose pretrial release must be decided by a judge. As a basis for this decision, a two-year recidivism rate is predicted by an ML model. 
However, when the ML model is trained on historical data, it tends to perpetuate historical discrimination such as racial profiling or, in some cases, racially discriminatory practices by prosecutors or judges. This means that a high recidivism rate may reflect historical discrimination rather than the individual's actual likelihood of committing a crime. Such an approach is called \myemph{bias preserving} by \shortciteA{wachter_bias_2021}, which includes approaches that optimize for ``classical fairML metrics'' such as predictive parity or equalized odds: While they try to equalize error rates in different subgroups as defined by PAs, they still aim to predict well in the -- potentially biased -- real world.
In contrast, we take a \myemph{bias transforming} approach by imagining a FiND world in which certain causal effects on recidivism that concern PAs are eliminated, using the prediction derived in that FiND world as the basis for our decision. This would underestimate the true 
likelihood of being convicted of a crime 
in the real world -- assuming that the real world is still characterized by racial discrimination -- but would more fairly reflect the individual’s likelihood of 
being convicted of 
a crime in an unbiased world. We will use this example throughout the paper.

\paragraph{College Admission} Consider a Black student who, because of structural inequalities, has attended underfunded schools with fewer educational opportunities, lived in a less safe neighborhood, suffered racial profiling, and so on. 
Suppose that the college admissions decision is based on a prediction of her probability of educational success based on, among other things, her SAT score.
%
A bias preserving approach would be based on training an ML model on
historical data, including biases that may have led to lower success rates for individuals from certain PA groups. This leads to several problems. First, the model might ``correctly'' (for the biased real world) predict further structural inequalities in college education, thus perpetuating these inequalities to the disadvantage of the individual. 
Furthermore, it might not be able to adequately predict the individual success rate, as it is likely that the Black student has greater potential than reflected by the SAT score because she achieved her score despite all the obstacles placed in her way. In our terms, this would mean that she 
actually has 
a higher task-specific merit (see definition below). 
And lastly, even if the person in question actually has a lower chance of success at college due to their poorer educational opportunities, the college could decide that the person should not have to take responsibility for this and still give the person the chance to make up for these deficits at university in order to break the circle of inequality.
Technically, we aim to overcome these problems by removing the direct and indirect effects of racialisation on her SAT scores. Within such an approach, it is possible to calculate her actual task-specific merit by removing only some path-specific effects while preserving others. 
As a result, she would have a higher SAT score in the FiND world, potentially leading to a higher predicted probability of success, justifying her admission to college.


\subsection{Related Work}

What exactly is understood by fairness is not disclosed by central laws \shortcite{clarke_text_2019,eu_eur-lex_2016} or statements from politics \shortcite{executive_office_2016}. Because of this, there exists a broad and now almost unmanageable body of literature on the topic of fairness in general and, in particular, on the fairness of ADM -- especially in the social sciences, law, and more recently in ML \shortcite<for overviews see, e.g.,>{agarwal_reductions_2018,barocas_fairness_2019,berk_fairness_2021,caton_fairness_2023,chouldechova_snapshot_2020,mehrabi_survey_2022}. 
However, most of their arguments are presuppositional and start a step further than we do here. When attempted, explanations of what fairness really means usually settle on vague definitions \shortcite{beck_kunstliche_2019,dator_chapter_2017,green_fair_2018,mainzer_fairness_2020,kleinberg_inherent_2017} -- notable exceptions being \shortciteA{loi_is_2022} and \shortciteA{kong_are_2022}.
A fundamental critique of statistical criteria as fairness measures such as ``false positive rate equality does not track anything about fairness, and thus sets an incoherent standard for evaluating the fairness of algorithms'' by \shortciteA{long_fairness_2021} is also formulated by \shortciteA{fleisher_algorithmic_2023,hedden_statistical_2021,loi_is_2022}, pointing out a special role of calibration.
Reflections on fairness as not (only) a technical challenge can be found in, e.g., \shortciteA{green_escaping_2022,lee_formalising_2021,selbst_fairness_2019,wong_democratizing_2020}.

There is a growing awareness in ML literature that basic assumptions that are not made explicit do indeed matter \shortcite{friedler_impossibility_2016,lee_fairness_2020,mitchell_prediction-based_2021,schwobel_long_2022}, but the reappraisal of these basic assumptions is still in its infancy.
Adding to the many concepts of \myemph{group fairness} \shortcite<see, e.g.,>[for an overview]{verma_fairness_2018}, most recently, the notion of \myemph{individual fairness}, highlighted in particular by \shortciteA{dwork_fairness_2012}, has resonated considerably \shortcite{bechavod_metric-free_2020,chouldechova_snapshot_2020,friedler_impossibility_2016,kusner_counterfactual_2017}.
This concept requires that similar individuals should be treated similarly and reflects a demand already comparably formulated by \shortciteA{guion_employment_1966} more than 50 years ago \shortcite<see also>[regarding the debate on test (un)fairness in the mid-20th century]{hutchinson_50_2019,thorndike_concepts_1971}.
For ranking tasks, \shortciteA{singh_fairness_2021} assume that ``[u]nfairness occurs when an agent with higher merit obtains a worse outcome than an agent with lower merit''.
\myemph{Individual counterfactual fairness} \shortcite{kusner_counterfactual_2017} goes in a different direction, stating that 
a fair decision exists if it turns out to be the same in the real world and in a fictitious world in which the individual in question belongs to a different protected group. 
This is arguably the work that is most related to our concept; however, after introducing our framework, we differentiate our proposal from that of \shortciteA{kusner_counterfactual_2017} in Section \ref{sec:comp_other_concepts}.
As described later, these definitions -- which take up causal concepts \shortcite<see also>{chiappa_path-specific_2019,chiappa_causal_2019,chikahara_learning_2021,grabowicz_marrying_2022,kasy_fairness_2021,kilbertus_avoiding_2017,kusner_long_2020} -- seem to produce useful results and are also supported by our considerations, but nevertheless do not represent essential aspects of the fairness concept. We note that our concept can be used to resolve confusions that have recently arisen in view of supposedly different definitions of fairness \shortcite{friedler_impossibility_2016,kusner_long_2020,kusner_counterfactual_2017,mitchell_prediction-based_2021}. 

To reveal the basic formal structure of the concept of fairness, it is not enough to go back to the great works of the 20th century, since these also proceed a step further and usually focus on which material criteria should be taken into account in the context of a just or fair distribution \shortcite{calsamiglia_decentralizing_2005}.
These considerations already build on the concept that we will present below 
\shortcite{rawls_theory_2003,roemer_theories_2018}.
For this reason, our considerations lend themselves equally well as a complement to the manageable ML literature that ties in with corresponding theories of distributive justice
\shortcite{binns_fairness_2018,gajane_formalizing_2017,kasy_fairness_2021,kuppler_distributive_2021}.

\section{The Basic Structure of Fairness}
\label{sec:defs}

The general understanding of fairness is regularly characterized as (i) typically concerning the treatment of people by people \shortcite{aristotle_nicomachean_2009,cambridge_dictionary_fairness,dator_chapter_2017,kleinberg_inherent_2017} and (ii) not being described as a concept, but merely by referring to normatively charged synonyms -- such as justice, equality, or absence of discrimination. At first glance, one might think that a more detailed definition of the term is superfluous and that fairness may be difficult to define, yet intuitively graspable. 
Suppose, e.g., that we have a cake from which two people are to receive a portion. It seems (initially) ``fair'' if each person gets one half of the cake. Fairness, then, is equated with ``equality'' or ``equal treatment''. But what if one person is starving and the other is well-fed and satiated? 
What if the cake is supposed to be a reward for a service previously rendered and one person has done twice as much as the other?

\subsection{Basis for Decision: Task-specific (In-)Equality}
\label{sec:equality}

In these considerations -- sometimes referred to as the difference between ``equality'' and ``equity'' -- 
lies the basic problem of arguments about fairness. These arguments always depend on the reference point of the evaluation: (1) Is it solely the distribution of the asset at hand? 
Or (2) should the point of reference also be the person concerned? In the second case, what is fair is determined by who is affected.

These rather trivial considerations can be translated into a theoretical framework, i.e., a \myemph{formal basic structure}. The fundamental aspects of this concept were already developed by Aristotle in his analysis of the nature of justice \shortcite{aristotle_nicomachean_2009} 
-- which is not considered by \shortciteA{lee_fairness_2020,lee_formalising_2021,zwick_philosophical_2016} in their approaches to invite Aristotle into the discussion about fairness --
and still provide a viable foundation for contemplating fairness today. Here, justice can be understood as mere adherence to the standards agreed upon in society (e.g., in laws) but also refers to the idea of equality. This idea of equality is the common root of what is meant by ``justice'' or ``fairness'' when used as a critical concept. Equality demands that \myemph{equals are to be treated equally and unequals are to be treated unequally}. In other words, if unequals are treated equally, this is unfair. If equals are treated unequally, this is also unfair.

Consequently, the decision-making basis for treatment is the question of whether or not people are equal. However, \myemph{equality is a strongly normatively loaded term}, because people possess infinitely many qualities and are therefore never exactly equal. In relation to certain situations, however, there is a normative stipulation that this difference between people should be irrelevant. 
Aristotle stated that this is particularly the case in private relations: if two people conclude a contract and it is a question of whether performance and consideration are balanced, it is irrelevant who these people are (e.g., the price for a bottle of water in a specific supermarket is $x$ -- no matter who buys it). Similarly, for the assessment of a penalty, it is irrelevant whether a rich person kills a poor person or vice versa. 
Because the decision on the treatment here can be made by means of a simple calculation (e.g., the price for two bottles of water is $2x$ -- no matter who buys it), Aristotle speaks in this respect of equality as \myemph{arithmetic or continuous proportionality}. 

In other situations, 
equality is said to depend on some characteristics of the people concerned; the relevance of the characteristics for the assessment of equality is decided normatively.  
Aristotle calls this the \myemph{worthiness} of people. We will also use the -- more contemporary -- term \myemph{task-specific merit} to refer to this concept of \myemph{task-specific equality}. For example, take the tax rate: usually, those who have a higher income also pay a higher tax rate, and vice versa. 
Because this kind of distribution decision must consider the balance of a more complex ratio, Aristotle speaks here of the \myemph{geometric or discrete proportionality}. The distribution ratio results in dependence on the task-specific merit negotiated in the political dispute: the ratio of the task-specific merit of person $i$ (e.g., the income) to the assets distributed to them (e.g., the tax rate) must correspond to the ratio of the task-specific merit of person $j$ to the assets distributed to them. 

As evidenced, \myemph{equality is always the result of a normative stipulation}. This is accompanied by an evaluation of what is to be brought into a relationship of equality -- only the things or assets that are distributed (arithmetic proportionality), or also the features of the people involved (geometric proportionality). 
For Aristotle, this depends on the subject area concerned: private dealings are decided by arithmetic proportionality, and government distribution is decided by geometric proportionality. 
Today, it is part of the political dispute whether private matters may be left in this sphere or whether state intervention according to the principles of the state distributive system is deemed necessary, e.g., when private companies discriminate racially. 
Due to the normativity of the concept of equality, the assessment of whether actions constitute equal or unequal treatment -- and are ``fair'' or ``unfair'' -- can vary widely depending on the system of norms involved. However, certain moments of consensus are now emerging, at least in certain regions of the world -- for example, with regard to the unequal treatment of women or ethnic groups. 

Nowadays, the so-called \myemph{Protected Attributes} (PAs) play a special role in the decision-making process, e.g., the characteristics listed in the US Civil Rights Act of 1964, in the Charter of Fundamental Rights of the European Union, or in Article 3 of the German Constitution. 
In some cases, the comparison and treatment of two people must not be based on these PAs (\myemph{PA-neutrality}). 
Since the attributes cannot act as causal factors -- if they are not to be considered for comparisons -- this decision is accompanied by the consideration of ignoring the consequences of these attributes as well, 
see e.g., the example of COMPAS \shortcite{angwin_machine_2016} which will be used in later sections of this work: if ethnicity has an influence on an offender's probability of recidivism, it seems natural to choose not to take this into account. 
Here, an underlying consideration may be that ethnicity is not the direct reason for the higher probability of recidivism, but rather that ethnicity has complex consequences for socialization processes, which then in turn have an effect on the higher recidivism probability, e.g., an average lower level of education or a certain place of residence. 
Another consideration might be that detection rates and conviction rates of criminal offenses might vary across ethnicities (e.g., due to racial profiling or biased law enforcement agencies), leading to higher records of recidivism rates.
Society may take responsibility for these consequences, wanting to keep them out of the decision-making process. However, 
because processes occurring in the life of an individual are usually not monocausal -- i.e., in the example, ethnicity is not the only causal factor for the level of education -- 
this is again a social negotiation process, at the end of which a normative decision is made as to who is to be attributed responsibility for which processes. 
These considerations make it evident that eventually a fictitious world massively corrected by normative evaluations becomes the basis for deciding on the treatments of individuals. 
In Section \ref{sec:ml} onwards, we will focus on the case of PA-neutrality, elaborate on how this fictitious world can be formalized and which consequences this has for the design of ML models that shall be used as components of ADM systems.

Conversely, there are constellations in which the PAs are specifically targeted 
in order to justify the inequality of people and, thus, their unequal treatment in the form of a preference for the feature bearers or a focus on their specific characteristics \shortcite<also called \myemph{PA-focus} or \myemph{affirmative action}, see, e.g.,>{coate_will_1993,foster_economic_1992,kalev_best_2006,keith_effects_1985}.
Examples could be (1) hiring policies that aim at achieving gender diversity or educational programs that actively target historically disadvantaged populations or (2) decisions on medical treatments that take into account, e.g., the physiological differences between people of different gender, also referred to as ``gender medicine'' \shortcite{baggio_gender_2013}.
The perspective depends on a normative decision as to whether the protection of the feature bearers in the respective 
task
is to be ensured ``only'' by means of exclusion of these features and their effects (PA-neutrality), or whether 
reality is to be actively reshaped according to certain objectives (PA-focus). 
In Aristotle's words, in the first case, the protected attribute must not be used to decide on task-specific merit, while in the second case, the same attribute is used to determine higher merit.
Thus, even the complex socio-political realities of the modern world fit into Aristotle's concept, while at the same time, it is clear that today complex normative decision-making processes are involved.
Even in situations where we conceive PAs, there is thus no generally fixed ``equal treatment'' or ``fairness'', but every concrete task demands answers to the above normative questions. 
As mentioned at the beginning, the specific, technical framework proposed in the remainder of this paper focuses on PA-neutrality, while a corresponding framework for PA-focus -- even if fitting into the same basic philosophical concept -- is left to future research.

\subsection{Decision Rule: Equal Treatment}

Once the normative basis for a decision has been established, the second step is to develop a decision rule that determines the extent to which the equality or inequality thus conceived is to be taken into account.
In the simpler case of arithmetic proportionality, a fair treatment does not depend on the task-specific merit:

\begin{definition}[Arithmetically fair treatment]
A treatment $\ti$ of an individual $i$ is called \myemph{arithmetically fair} if and only if it is the same as for any other individual, i.e., does not depend on an individual's task-specific merit $\wi$
$$\ti = k \quad \forall i \in \{1, \dots, n\}, k \in \mathbb{R},$$
where $n \in \mathbb{N}$ is the number of individuals in the entire population.
\end{definition}

In the case of geometric proportionality, it must hold that the ratio of treatment $t$ to task-specific merit $\w$ is the same for any comparison of two individuals $i$ and $j$, i.e., 
$$\frac{\ti}{\wi} = \frac{\tj}{\wj} = k \quad \forall i,j \in \{1, \dots, n\}, k \in \mathbb{R}.$$ 
In other words, we can define a treatment function $s: \W \rightarrow T, \ \w \mapsto t$, where 
typically $\W \subseteq \mathbb{R}$ and 
$T \subseteq \mathbb{R}$:

\begin{definition}[Geometrically fair treatment]
A treatment $\ti$ of an individual $i$ is called \myemph{geometrically fair} if and only if it is a linear function of the individual's task-specific merit $\wi$, i.e., 
$$\ti = s(\wi) = k \cdot \wi \quad \forall i \in \{1, \dots, n\}, k \in \mathbb{R}.$$ 
\end{definition}

The task-specific merit $\wi$ may be directly identified with some observable feature $\vi$ (such as the income in the tax rate example), but it may also be defined as a latent and non-observable feature $\zi$.
Finally, it may be a combination of observable and non-observable features, i.e., $\wi = f(\vi, \zi)$, where $f(\cdot)$ is another normative function.
If PAs are present, the effect of these features and hence the task-specific merit of an individual may be modified by the PAs, referring to their counterparts in a fictitious, normatively desired world, see Section \ref{sec:ml} for details.
In the example of pretrial decisions -- such as in the COMPAS example introduced shortly in this paper -- $\vi$ and $\zi$ may be the type of crime and probability of recidivism, respectively.
The duty of ML in the ADM process will be to provide an estimate of the non-observable feature(s) $\zi$, where their meaning is possibly modified by the PAs. These estimates serve as the basis for assessing the task-specific equality of individuals, i.e., for deriving their task-specific merit $\wi$, which in turn is the basis for the decision of their treatment $\ti$ via $s(\wi)$, see below.

It may be decided to modify the treatment function $\sfu$ 
in a more flexible way, e.g., if a tax rate $\sfu$ is raised with higher income $\wi$, but this is done step-wise rather than continuously -- and above a certain income, not at all. The function $\sfu$ can thus be normatively corrected:

\begin{definition}[Modified fair treatment]
\label{def:mfair}
A treatment $\ti$ of an individual $i$ is called modified \myemph{fair} if and only if it is determined by a monotonic\footnote{Note that we do not require the function $\sfu$ to be \textit{strictly} monotonic in general; the function is allowed to have plateaus, such as in the tax example. The concrete form of the function and the concrete value of $k$ are normative choices.} function of the individual's task-specific merit $\wi$, i.e., 
\begin{equation*}
\ti=s(\wi) \quad \forall i \in \{1, \dots, n\}.
\end{equation*}
\end{definition}

This approach still allows for (i) equal treatment of equals 
and (ii) unequal treatment of unequals in a normatively defined way that is then considered to be fair. 
To enhance readability, the term ``fair treatment'' refers to this notion in the remainder of the paper, if not stated otherwise.
Even if it would be mathematically straightforward to allow for $\sfu$ to be more flexible, a non-monotonic $\sfu$ would not be in line with the above philosophical concept. One will demand 
strict monotonicity if unequals are always to be treated unequally. Geometrically and arithmetically fair treatments result from special choices of $\sfu$.

Thus, again, valuation decisions arise that add a normative dimension, making it three normative questions in this scenario of Def. \ref{def:mfair}: 
(1) How is the task-specific merit $\wi$, i.e., the measure of task-specific equality of individuals, defined? (2) Which attributes are defined as PAs (if there are any)?
(3) Shall the treatment function $s(\wi)$ be modified, and if so, how?  
%

\newpage
\section{Role of ML in the ADM Process}
\label{sec:ml}

Now that we have defined the general concept of fairness, we can turn to answer the questions of to what extent an ML model used in the context of an ADM process can contribute to unfair treatment and of what this entails for the design of ML models. First, in Section \ref{sec:adm}, we review the steps of an ADM process with respect to the applicability of the notion of fairness and illustrate it using the COMPAS example -- focusing on the goal of PA-neutrality rather than PA-focus (see again Section \ref{sec:equality}). In Sections \ref{sec:ml_ohne_pa} and \ref{sec:ml_mit_pa} we discuss the contributions of the ML model and propose concrete solutions for the situations without and with PAs, respectively. In Section \ref{sec:comp_other_concepts} we compare our framework with two seemingly related concepts from the literature. After this core methodological part, we illustrate its application in Section \ref{sec:implementation}.

\subsection{Fairness in the ADM Process}
\label{sec:adm}

\begin{figure}[ht]
    \centering
    \includegraphics[width=\textwidth]{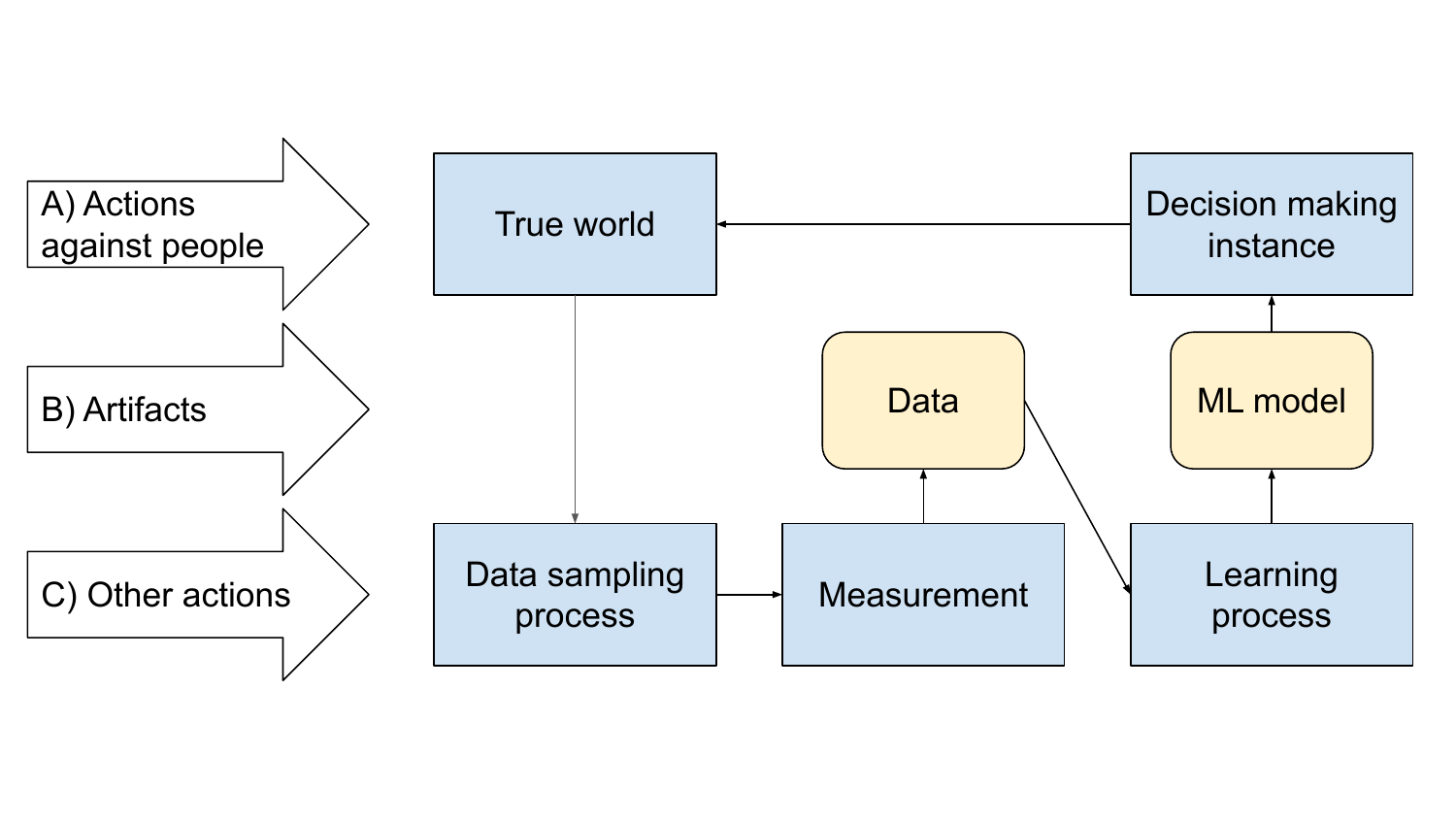}
    \caption{Simplified ADM process, following \shortciteA{suresh_framework_2021}.}
    \label{fig:world_actions}
\end{figure}    

The ADM process is explained in detail in \shortciteA{suresh_framework_2021}, while we consider a reduced version 
(see Figure \ref{fig:world_actions}).
We divide the process into three categories: (A) actions against people, (B) artifacts, and (C) other actions. As shown in Section \ref{sec:defs}, the notion of fairness is applicable only to category (A) and not directly applicable to the ML model.
However, the results of categories (B) and (C) affect the estimated ``task-specific merit'', i.e., the measure of task-specific equality of individuals that is the basis of decisions for actions against people.
As mentioned above, the task-specific merit $\wi$ may consist of observable features $\vi$ (like the type of crime) and non-observable features $\zi$ (like the recidivism probability), i.e., $\wi = f(\vi, \zi)$. For ease of presentation, we will ignore any observable features in the following and set $\wi = \zi$.
For a classification task, $\zi$ is the individual probability of success $\mathbb{P}(y = 1 | i)=\pii$ of an event $y \in \{0,1\}$; for a regression task, it is the individual expected value $\mathbb{E}(y|i) = \mui$. Thus, indirectly, the elements of categories (B) and (C) can induce unfair actions. 
We will hold to the classification scenario in the following for notational clarity, but switching to the regression -- or survival -- scenario is straightforward, thereby remedying a critique towards fairML's focus on classification formulated by \shortciteA{hutchinson_50_2019}.
Before we shed more light on the role of ML in this process (Sections \ref{sec:ml_ohne_pa} and \ref{sec:ml_mit_pa}), 
we 
highlight the crucial difference between the ML model and actions based on the model 
as well as analyze the effect of the occurrence of PAs.

In the much-cited COMPAS example \shortcite{angwin_machine_2016}, the  
individual probability $\pii$ of recidivism within two years ($y \in \{0,1\}$) \shortcite{larson_how_2016} is the basis for deciding how to treat defendant $i$. Since $\pii$ is unknown, an ML model $\pihf$ 
is used to estimate this probability, based on a feature vector $\xi$.
The goal of the ML model \shortcite<see, e.g., >{pedreshi_discrimination-aware_2008} is to assign different recidivism probabilities as accurately as possible to different individuals, based on the observed features: 

\begin{definition}[Statistically discriminative model w.r.t.\ feature  \textit{X}] 
An ML model $\pihf$ is called \myemph{statistically discriminative w.r.t.\  feature $X$} if there is at least one pair of individuals $i$ and $j$ 
who differ only with respect to feature $X$ and are assigned different predicted probabilities $\pihi \ne \pihj$.\footnote{Note that from here on, individuals $i$ and $j$ are not limited to be members of a finite training set but are assumed to be members of a potentially infinite population, i.e., their feature vectors can be any point in the feature space.}
\end{definition}

This is either the case or not the case for each feature. However, a fairness evaluation can only be applied to the action based on this evaluation. From this, we derive similar to \shortciteA{guion_employment_1966}:

\begin{definition}[Descriptively unfair treatment] 
Assume a pair of individuals $i$ and $j$ who differ only with respect to feature $X$. Assume that feature $X$ is not a causal reason for a difference in the true probabilities, i.e., 
$$\pii = \pij.$$
A treatment is called \myemph{descriptively unfair w.r.t.\  feature $X$} if these individuals are treated differently, i.e., 
$$\ti (= s(\pihi)) \neq \tj (= s(\pihj)),$$ 
in a process due to differing estimated individual probabilities $\pihi \ne \pihj$.
\end{definition}

\begin{example}
Assuming that the recidivism probability $\pi$ does not causally depend on the ethnicity $X$, yet two persons who differ only with respect to ethnicity (i.e., have the same true recidivism probability) would be assumed by the ML model to have different recidivism probabilities, and therefore there would be different judicial decisions, then this unequal treatment would be descriptively unfair -- regardless of whether ethnicity is a PA or not, because equals (equal task-specific merit $\pii = \pij$) would be treated unequally (unequal treatment $\ti \neq \tj$).
\end{example}

If, on the other hand, $X$ is causal for a difference in $\pi$, i.e., $\pii \neq \pij$, then a differing decision basis due to $X$, i.e., $\pihi \neq \pihj$, and a resulting difference in treatment, i.e., $\ti  \neq \tj $, cannot be said to be descriptively unfair; unequals are treated unequally.
This evaluation may change with the introduction of PAs, as derived in Section \ref{sec:defs} and similarly noted by \shortciteA{lee_fairness_2020}, as the assessment of task-specific equality changes:

\begin{definition}[Normatively unfair treatment]
\label{def:norm_unfair_treat}
Assume a pair of individuals $i$ and $j$ who differ only with respect to feature $A$. Assume that feature $A$ is a causal reason for a difference in the true probabilities, i.e., 
$$\pii \neq \pij.$$
Assume that feature $A$ is a PA.
A treatment is called \myemph{normatively unfair w.r.t.\  feature $A$} if these individuals are treated differently, i.e., 
$$\ti (= s(\pihi)) \neq \tj (= s(\pihj))$$
in a process due to differing estimated individual probabilities $\pihi \ne \pihj$, as feature $A$ must not be invoked 
for the determination of task-specific equality. 
\end{definition}

\begin{example}
Suppose that the probability of recidivism $\pi$ depends causally on the ethnicity $A$, but the judicial decision should not -- for normative reasons -- depend on differences in $\pi$ due to ethnicity, e.g., 
because society decides not to let the individual bear the responsibility for the grievance that ethnicity is causal for the probability of recidivism (e.g., via racial profiling and higher re-arrest rates), but to take it from them and bear it as a whole society.
\end{example}

So, the measure of task-specific equality is no longer the true recidivism probability $\pi$ in the real world, but the true recidivism probability $\psi$ in a fictitious, normatively desired (FiND) world:

\begin{definition}[FiND world] 
A FiND world is a fictitious, normatively desired world, where the PAs have no causal effects on the target variable, neither directly nor indirectly, i.e., $\psii = \psij$ for the pair considered in Definition \ref{def:norm_unfair_treat}. 
\end{definition}

This mirrors the demand of several laws to not differentiate between individuals based on their PA (e.g., gender, ethnicity), which means that individuals are to be considered as being equal if they only differ in their PA. 
This also means that subsequent effects of their PA on other features must be eliminated, therefore removing all direct and indirect effects of the PA on the target to reach the FiND world (i.e., remove dashed arrows in the Directed Acyclic Graph (DAG) shown in Figure \ref{fig:causal_graph}). 

In debates such as the discussion surrounding COMPAS, a fairly broad consensus has been reached that combating discrimination means not only tackling direct discrimination but also indirect discrimination through proxy variables. 
However, several recent papers in fairML  \shortcite<e.g.,>{chiappa_path-specific_2019,nabi_fair_2018,nabi_learning_2019,pan_explaining_2021} argue for removing only some path-specific effects where the path itself is deemed unfair.
While in some cases, indirect discrimination might be allowed, 
\shortciteA{wachter_affinity_2019} argues that ``[i]ndirect discrimination is only lawful if a legitimate aim is pursued and the measure is necessary and proportionate'' and describes examples of European case law where this issue has been debated, concluding that the hurdles for justifying indirect discrimination are rather high. 
As there may therefore be cases where some paths starting from the PA are not considered unfair, we need methods that deal with path-specific effects. 
Our methodology is easily extended to such admissible path-specific effects: Let $X_{ps}$ be a feature where the path from the PAs through $X_{ps}$ to the target is normatively defined as not being unfair. Then the effects of the PAs on this $X_{ps}$ are not removed in the definition of the FiND world (and hence the corresponding arrows in Figure \ref{fig:causal_graph} remain solid):

\begin{definition}[FiND world -- admissible path-specific effects]
\label{def:find-ps}
A FiND world with admissible path-specific effects is a fictitious, normatively desired world, where the PAs have no causal effects on the target variable, neither directly nor indirectly -- with the exception of path-specific effects going through some features $X_{ps}$. 
\end{definition}

The question of which paths should be removed, and thus the concrete definition of the FiND world, is a purely normative and potentially highly controversial issue. Our framework covers a wide spectrum of potential answers to these normative questions.

Note that it can be argued that there are several FiND worlds that fit either of the two definitions above. 
The concrete definition of the FiND world depends on the specific actions that are carried out to remove the dashed arrows in the DAG. This relates to the definition of what exactly the baseline ``without PA effects'' is. 
As an example, let us assume that \textit{ethnicity} is a binary PA with values \textit{Non-White} and \textit{White}. For mapping the number of priors ($X_P$) into a world without effects of \textit{ethnicity}, should we remove the effect of belonging to the \textit{White} subgroup (as compared to the \textit{Non-White} subgroup) on $X_P$? This would make the \textit{Non-White} subgroup the baseline. Or should we remove the effect of belonging to the \textit{Non-White} subgroup (as compared to the \textit{White} subgroup) on $X_P$? This would make the \textit{White} subgroup the baseline. 
Alternatively, the baseline could be something in between, like the marginal distribution of $X_P$. 
The choice of baseline will change the concrete values predicted by an ML model for that world. 
However, all such baselines are equally valid when aiming at fair, i.e., PA-neutral, treatment. In the remainder, we will assume that a decision on the baseline has been made and work with the such defined FiND world.

Once we have moved to this world after defining the PAs, we use $\psii$ (instead of $\pii$) as a measure of task-specific equality, i.e., introducing PAs has changed the definition of task-specific merit in a specific use-case normatively; 
hence, a fair treatment would result by $\ti = s(\psii)$. The role of ML is to estimate $\pii$ or $\psii$ accurately, respectively. 

\begin{figure}[th]
     \centering

\begin{tikzpicture}[x=12in, y=5in]
\node[ellipse, draw] (v0) at (0.564,-0.408) {$\boldsymbol{C}$};
\node[ellipse, draw] (v1) at (0.564,-0.543) {$X_D$};
\node[ellipse, draw] (v2) at (0.564,-0.678) {$A$};
\node[ellipse, draw] (v3) at (0.779,-0.543) {$Y$};
\node[ellipse, draw] (v4) at (0.349,-0.543) {$X_P$};
\draw [->] (v0) edge (v1);
\draw [->] (v0) edge (v3);
\draw [->] (v0) edge (v4);
\draw [->] (v1) edge (v3);
\draw [dashed, ->] (v2) edge (v4);
\draw [->] (v4) .. controls (0.473,-0.261) and (0.646,-0.261) .. (v3);
\draw [dashed, ->] (v2) edge (v1);
\draw [dashed, ->] (v2) edge (v3);
\end{tikzpicture}
    \caption{Illustrative DAGs for the COMPAS example. In the FiND world, only solid arrows exist, and in the real world, all arrows exist. Features are \textit{ethnicity} (protected attribute $A$ with classes \textit{Non-White} and \textit{White}), \textit{age, gender} (confounder $\boldsymbol{C}$), \textit{number of priors} ($X_P$), and \textit{charge degree} ($X_D$). The target is \textit{recidivism in two years} ($Y$).}
   \label{fig:causal_graph}
\end{figure}

\subsection{Contribution of ML -- without PAs}
\label{sec:ml_ohne_pa}

We first consider the situation where no PA is present and show that, even then, the use of an ML model can induce fairness problems. In Section \ref{sec:ml_mit_pa}, we turn to the situation where PAs are present.

\subsubsection{Can ML induce unfairness, and if so, how?}

In Section \ref{sec:defs}, we observed that the treatment $\ti$ of an individual $i$ based on task-specific merit $\wi$ is fair if and only if it follows the normative decision rule $\sfu$, i.e., if and only if $\ti = s(\wi)$. 
In Section \ref{sec:adm}, we observed that in the context of an ADM process, the task-specific merit $\wi$ often corresponds to the individual probability $\pii = \mathbb{P}(y = 1|i)$. 
Since the true individual probability $\pii$ is unknown in practice, we use data to estimate $\pii$. 
In doing so, two key steps introduce imprecision that can induce unfair treatment and are described below (see also Table \ref{tab:two_worlds}). 

\begin{table}[ht]
  \centering
  \begin{tabular}{lccc}
    \toprule
    &Real world & Warped world & FiND world\\
    \midrule
    True target probability & $\pii$ & $\piti$ & $\psii$ \\
    Fair treatment & $\spii$  &  $\spiti$ & $\spsii$  \\
    Treatment w/ coarse inf. & $\spixii$& $\spitxti$ & $\spsixii$ \\
    Treatment w/ ML model&$\spixih$&$\spitxtih$ & $\spsixih$\\
    \bottomrule
  \end{tabular}
   \caption{Coarse information due to finite feature space $\Xspace$ and estimation via ML introduce errors. Point of reference changes with presence of PAs (columns, see also Section \ref{sec:ml_mit_pa}).}
   \label{tab:two_worlds}
\end{table}


\paragraph{Coarsening of Information}
The first step is to coarsen the information by basing the treatment not on the individual probability $\pii$ but on a group probability $\pixii$ that assigns the same value to all individuals $I_\xv = \{i: \xi = \xv\}$ with the same combination of observed features $\xv$.
Naturally, the function $\pi: \Xspace \rightarrow [0,1]$ is as true and unknown as the individual probabilities $\pii$. For any feature combination $\xv$, this function is the best possible approximation of the different $\pii$ of the different individuals $I_\xv$ sharing that feature combination, 
based on the available $p$ features. 

This coarsening of information introduces fairness problems, since individuals $I_\xv$ are treated the same even though they are -- except for the features collected -- not the same. With the exception of degenerate special cases where all individuals in a group $I_\xv$ are identical or for a few individuals where $\pii = \pixii$ happens to hold, this results in 
$$\pii \ne \pixii \Rightarrow \spii \ne \spixii = \ti \ \forall i \in \{1, \dots, n \}$$
(if $\sfu$ is injective), so (almost) all individuals are treated unfairly -- even if $\pif$ is known.

This step is only appreciated in a few works, such as \shortciteA{friedler_impossibility_2016,kleinberg_inherent_2017,mitchell_prediction-based_2021,singh_fairness_2021}. However, it is very important for fairness considerations of ADM systems, since this means that already reducing the information regarding an individual to finitely many features is a gateway to unfairness -- even if we knew the true $\pif$, 
even before introducing PAs, and even before estimating an ML model.

\paragraph{Estimation by ML}
Estimating the unknown function $\pif$ by $\pihf$ introduces two types of error: The \myemph{estimation error} comes from the fact that the learner has only a finite amount of training data $\D = \Dset$ available. This error converges to $0$ for $n \rightarrow \infty$.
Should the true $\pif$ not be part of the hypothesis space $\Hspace$, a non-reducible \myemph{approximation error} remains. 

\subsubsection{Evaluation regarding fairness}
\label{sec:ml_ohne_pa_eval}

Evaluation of the ML model $\pihf$ can be done at two levels, namely (i) with respect to $\pif$ or (ii) with respect to $\pii$. 
Considering level (i), for cases in which every conceivable feature combination $\xv \in \Xspace$ is represented sufficiently often (only possible in the case of few categorical features), the mean observed and predicted probabilities can be compared directly -- for example, via the L2 norm of the differences of the group means.
This value should be as small as possible on a test set. (Note, however, that using the L2 norm is another normative choice.)
Since this (a) is only conceivable for special data situations and (b) ignores the imprecision introduced by coarsening $\pii$ via $\pixii$, 
an evaluation with respect directly to 
$\pii$ seems more purposeful: 
We recall that a treatment $\ti$ is fair if and only if $\ti = \spii$, so here, if and only if $\spixih = \spii$. 
Thus, a sufficient condition for fair treatment is $\pixih = \pii$, 
which can be seen as an individual version of well-calibration \shortcite{kleinberg_inherent_2017,verma_fairness_2018}:

\begin{definition}[Individually well-calibrated model] 
An ML model $\pihf$ is called \myemph{individually well-calibrated for individual} $i$ if $\pixih = \pii$; the model is called \myemph{individually well-calibrated} if $\pixih = \pii \quad \forall i \in \{1, \dots, n\}$.
\end{definition}

Note that a direct generalization to regression, survival tasks, etc., is possible by replacing the probabilities with expected values. For a strictly monotonic function $\sfu$, this is also a necessary condition for fairness, i.e., a treatment $\spixih$ is to be called unfair if the model is not individually well-calibrated for individual $i$. 

Although the notion of fairness refers to actions against individuals, empirical evaluation cannot be performed individually; evaluation is only possible by considering appropriate groups. This poses a massive problem, since any group definition runs the risk of assigning individuals to a group that is inappropriate with respect to their true probability $\pii$. In particular, it falls short to define a group based only on a single feature. Rather, it is necessary to consider all computationally identifiable subgroups as is done in, e.g., \shortciteA{friedler_impossibility_2016,hebert-johnson_multicalibration_2018,kearns_preventing_2018,kim_multiaccuracy_2019}, and check well-calibration in all these subgroups.
Since exact equality $\pixih = \pii$ is not achievable in practice, a pragmatic tolerance range should be conceived. The above equalities in the definitions of individual well-calibration are then relaxed to $\pixih \in (\pii - \epsilon, \pii + \epsilon)$ with a tolerated deviation of $\epsilon \in \mathbb{R}^+$.

Here, also the group fairness metrics such as \myemph{equalized odds}, \myemph{predictive parity}, etc. \shortcite<see, e.g.,>{verma_fairness_2018} have their value: We can think of them as \myemph{fairness-related performance metrics} that allow us to obtain more nuanced information on the model performance in subgroups as this goes beyond well-calibration. Also here, the subgroups are not defined by PAs -- since there are no PAs in this scenario -- but all computationally identifiable subgroups should be considered for a holistic view of the model's performance.

\subsection{Contribution of ML -- with PAs}
\label{sec:ml_mit_pa}
We have observed that even without the presence of PAs, the use of empirical methods 
can induce fairness problems. We now analyze what changes 
through the introduction of PAs. First, we elaborate on the FiND world and its approximation via warping, and then we turn to the question of evaluating the model regarding fairness.

\subsubsection{FiND world and approximation via warping}
\label{sec:ml_mit_pa_fair}

In Section \ref{sec:adm}, it was shown that in the presence of a PA, the basis for decisions must be PA-neutral to be able to achieve fair treatment, i.e., the normative decision can be made to use the PA-neutral true probability $\psii$ instead of $\pii$ as a measure of task-specific equality. This $\psii$ describes for an individual $i$ the probability $\mathbb{P}(\yfind = 1 | i)$ for an 
event $\yfind \in \{0,1\}$ in a FiND world in which the PA has no causal effect on this event, neither directly nor indirectly.\footnote{While in the following we assume that all PA effects are removed for the definition of the FiND world, the normative decision can be made not to remove some admissible path-specific effects, see Definition \ref{def:find-ps}. The following methodology also applies in this case, we just need to define some FiND world normatively.} 
To differentiate the target in the FiND world from the originally observed event $y$ in the real world, we denote it as $\yfind$.
As above, treatment $\ti$ of an individual $i$ is then said to be fair if and only if it follows the normatively specified decision rule $\spsii$, with changed decision basis $\psii$. 
Thus, in the example, we consider the individual recidivism probability $\psii$, 
\myemph{in a fictitious world where the PA has no causal effect on recidivism}.

Since the PA is not supposed to have a causal effect on 
$\yfind$, 
the normative decision can be made to exclude certain indirect effects of the PA on $\yfind$ via other features. 
Therefore, \myemph{the feature vector of individual $i$ is also to be considered PA-neutral}, i.e., $\xfindi$. This does not mean that the PA is removed  \shortcite<hence, goes beyond ``fairness through unawareness'', see >{gajane_formalizing_2017}. 
Rather, only those influences that (possibly via detours) affect $\yfind$ are removed (see dashed arrows in Figure \ref{fig:causal_graph}). 
In the COMPAS example, individuals might have not only a different recidivism probability when the effect of the PA is removed but also, e.g., a different income or residence, assuming that in the real world, the PA also has a causal effect on these features.

It is also not certain that the relation of event and features is identical in the real and the FiND world after potentially causal effects are removed, so the function $\psif$ we are looking for is also potentially different from $\pif$.
In the COMPAS example, the influence of features such as income and residence on the recidivism probability might be different in the 
two worlds.
Thus, we are confronted with a different situation than in Section \ref{sec:ml_ohne_pa}, since we want to learn contexts in a fictitious world, with a different data-generating process 
(see also Table \ref{tab:two_worlds}, last column).

Since we never have access to FiND-world data in an empirical use case, we must approximate this data. We dub this approximation \textit{warping}, which results in data from the \textit{warped world}. Targets and features in this world are denoted $\yti$ and $\xti$, respectively, and the true target probability in the warped world is $\piti = \mathbb{P}(\yt = 1 | i)$. 
In Section \ref{sec:implementation}, we discuss and experimentally compare two concrete warping methods; for the remainder of this section, we simply assume that such a warping exists.

\subsubsection{Evaluation regarding fairness}
\label{sec:ml_mit_pa_eval}

We assume that -- based on warped-world data $\Dt = \Dsett$ -- we have trained a model $\pitifh$. 
For an individual $i$, we obtain an estimate $\pitxtih$, i.e., an approximation of its warped-world probability $\piti$ by warped features $\xti$.
Naturally, we introduce imprecision analogously as described above by coarsening information and estimation by ML, but this time in the warped world (see also Table \ref{tab:two_worlds}, second column). 
Per definition, the treatment $\ti = \spitxtih$ is fair if and only if $\spitxtih = \spsii$.
Thus, 
a sufficient condition for fair treatment includes the FiND world probability $\psii$:

\begin{proposition}
A treatment $\ti = \spitxtih$ of individual $i$ is fair if $\pitxtih = \psii$. For a strictly monotonic $\sfu$, this condition is also necessary for fair treatment.
\end{proposition}

\textit{Proof:} By definition, the treatment $\ti$ is fair if and only if $\ti = \spsii$.

``$\Leftarrow$'', i.e. ``$\ti$ is fair if $\pitxtih = \psii$'':
\begin{eqnarray*}
   \pitxtih &=& \psii \\
   \Rightarrow \spitxtih &=& \spsii\\
   \Rightarrow \ti &=& \spsii
\end{eqnarray*}

``$\Rightarrow$'', i.e. ``if $\ti$ is fair, then $\pitxtih = \psii$'':
\begin{eqnarray*}
   \ti &=& \spsii\\
   \Rightarrow \spitxtih &=& \spsii\\
   \stackrel{s() \ \text{str. mon.}}{\Rightarrow} \pitxtih &=& \psii \quad \qed
\end{eqnarray*}

We can only test this condition 
with access to $\psii$, which we do not have in practical use cases. However, we can divide the problem into two independent tasks: 

\textit{(1) -- Warping:} We have to find a warping such that the warped world approximates the FiND world as well as possible, i.e., that the structural causal models \shortcite<SCM, see>{pearl_causality_2009} in both worlds are similar and that the corresponding joint probability distribution $\tilde{\mathbb{P}}$ in the warped world approximates the joint probability distribution $\mathbb{P}_F$ in the FiND world. This means that warped-world data $\Dt$ can approximately be seen as a sample from the FiND world data-generating process $\mathbb{P}_F$ and that the individual observations approximate their FiND world counterparts, i.e., $\xti \approx \xfindi$ and $\yti \approx \yfindi$. Consequently, the warped world probabilities should also approximate their FiND world counterparts, i.e., $\piti \approx \psii$. 

\textit{(2) -- ML model:} With data from that warped world, we can train a model that optimizes predictive performance -- as usually done in ML and without the need to account for PAs in any special sense (since this was already accounted for in the warping) and also without the need for fairness-specific evaluation metrics. 
Here we aim at a version of individual well-calibration, analogously as described in Section \ref{sec:ml_ohne_pa_eval}, but now concerning the warped world probabilities $\piti$. We could think of this as being in the situation without PAs, and now we only have to make sure that the ML model predicts as accurately as possible. Hence, the evaluation of task (2) is straightforward and should be done as described in Section \ref{sec:ml_ohne_pa_eval}, i.e., considering all computationally identifiable subgroups and possibly making use of fairness-related performance metrics.

The better the warping (1) and the better the performance of the model (2), the closer we get to fair treatments:
\begin{equation*}
    \ti = \spitxtih \stackrel{(1)}{\approx} s(\pitilh(\xfindi)) \stackrel{(2)}{\approx} s(\piti)  \stackrel{(1)}{\approx} s(\psii)
\end{equation*}

This means the evaluation of the warping (1) is disentangled from the evaluation of the prediction model (2) in the warped world. 
However, concerning task (1), we still face the complication that we do not know the true FiND world distribution $\mathbb{P}_F$ nor have access to FiND world data to compare warped world data against. 
Evaluations should hence focus on the validity of the warping method itself, which includes (at least) two parts: (i) the causal effects on the paths descending from the PAs have to be estimated correctly, (ii) the warping method has to ensure that it only corrects for these causal effects of the PAs and that other influences on individual observations remain untouched.
Another important point to keep in mind is that for now we assumed that the perceived DAG in the real world is correct. However, in practice we will face the challenge of deciding for a DAG based on expert knowledge and analytical tools such as causal discovery. This step brings along a certain amount of uncertainty that should be accounted for.
We leave it to future work to develop concrete measures for the validity of a warping method and for incorporating the uncertainty regarding the true real-world DAG. 
In the illustrative experiments below, we assume that the DAG is correct and that the warping is valid. We then focus on descriptive evaluations, comparing real-world and warped-world predictions for individuals, identifying the most discriminated individuals, and comparing these values on the level of PA-groups.




\subsection{Comparison to Other Concepts}
\label{sec:comp_other_concepts}

In this section, we compare our proposed framework with two (seemingly) related notions that have been previously discussed in the literature: The legal concept of \myemph{substantive equality} in EU non-discrimination law together with the definition of \myemph{bias transforming fairness metrics} as introduced by \shortciteA{wachter_bias_2021} and \myemph{counterfactual fairness} by \shortciteA{kusner_counterfactual_2017}.

\paragraph{Relation to Substantive Equality and Bias Transforming Fairness Metrics}

\shortciteA{wachter_bias_2021} discuss the ``legality of fairness metrics under EU non-discrimination law'' and distinguish two normative concepts described in the EU non-discrimination law, namely \myemph{formal equality} and \myemph{substantive equality}. While in formal equality, the aim is ``to not make society more unequal than the status quo'', in substantive equality ``true equality can only be achieved by accounting for historical inequalities which actively ought to be eroded''. Building on this, they define concepts of \myemph{bias preserving fairness metrics} -- that aim at formal equality -- and \myemph{bias transforming fairness metrics} -- that aim at substantive equality. Most interestingly, this can be related to our concept and the cases of no PAs (in Section \ref{sec:ml_ohne_pa}) and with PAs (in Section \ref{sec:ml_mit_pa}), respectively: If no PAs are present, we aim at individual well-calibration which basically means that the ML model shall reproduce the true data-generating process as well as possible, i.e., perpetuating the status quo. However, if PAs are present, we account for historical inequalities -- by moving to the FiND world -- and actively erode them -- by using warped world data for training and prediction. While our concept is essentially philosophically backed up by \shortciteA{wachter_bias_2021}, we contribute by translating this idea into an actionable framework for designing ML models. 

\begin{figure}[ht]
     \centering
         \includegraphics[width=\textwidth, trim=15 155 20 150, clip]{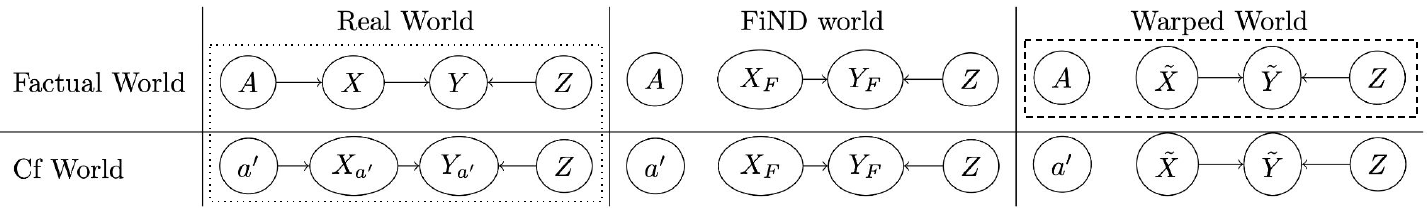}
         \caption{Relating our concept to counterfactual (Cf) fairness by \shortciteA{kusner_counterfactual_2017} on DAGs comprising PAs ($A$), features descending from $A$ ($X$), other features ($Z$), and a target $Y$. We argue to train, evaluate, and predict in the warped world (dashed box), while Cf fairness compares predictions in the factual world and the Cf world (dotted box).}
         \label{fig:cf_comparison}
\end{figure}

\paragraph{Relation to Counterfactual Fairness}
Counterfactual (Cf) fairness \shortcite{kusner_counterfactual_2017} also considers a fictitious world in comparison to the real world. While seemingly related at first glance, their concept differs substantially from our concept of FiND and warped worlds; see Figure \ref{fig:cf_comparison} for a comparison with a very simple DAG, where the top left DAG shows the factual, real world.
In our FiND world (top middle DAG), the PAs have no causal effect on the target and also not on features mediating this effect. The individual still belongs to its original class of the PA but with changed real-world descendants of the PA. In their Cf world (bottom left DAG), an individual belongs to a different class of the PA, which also affects descendants.
This means the FiND world changes (some) features and the target by
modifying causal relationships regarding the PA, while the Cf world 
intervenes on the PA, 
maintaining real-world causal relationships. 
We then approximate the FiND world through the warped world, train an ML model, evaluate the model, and predict new observations with the model in the warped world (dashed box). Cf fairness trains an ML model in the factual, real world, and evaluates it by comparing its predictions using factual and counterfactual values for the PA and remaining features 
to assess fairness (dotted box).

Despite these differences, we fulfill Cf fairness in a certain sense: Consider a Non-White person $i$. Cf fairness demands that its prediction (using its factual feature vector $\xi$) 
is the same as for its counterfactual $j$, where the person belongs to class White, i.e., 
using $\xv^{(j)} = \xv^{(i)}_{a'}$ ($\neq \xi$ in general). 
Since these two fictitious persons have the same task-specific merit and hence the same FiND-world feature vectors $\xfindi = \xfindj$ in our concept, we demand that they have the same warped world feature vectors $\xtil^{(i)} = \xtil^{(j)}$, which leads to equal predictions $\pitilh(\xtil^{(i)}) = \pitilh(\xtil^{(j)})$. 
In this world, the PA has no effect, which means that $\xtil$ does not contain descendants of the PA. With Lemma 1 of \shortciteA{kusner_counterfactual_2017}, this fulfills Cf fairness. However, as shown above, this is just a necessary condition for fairness and  
a form of individual well-calibration is paramount for achieving fairness. 
Measuring predictive performance must take place in the ``correct'' world, i.e., in the real world (if no PAs are present), or in the FiND world -- approximated by the warped world -- (if PAs are present).

\section{Implementation of the Concept}
\label{sec:implementation}

In this section, we tie together what was introduced so far and illustrate how our framework could be used for tackling fairness issues in an applied use case. Section \ref{sec:FiND-algo} outlines how to assess the FiND world via warping and how to train and predict in the warped world. Section \ref{sec:experiments} presents illustrative experiments with the COMPAS data. In the Appendix, Section \ref{sec:appl-usecase} describes a workflow for an applied ADM use case.

\subsection{Warping via Rank-Preserving Interventional Distributions}
\label{sec:FiND-algo}

We have observed so far that for valid training of a model as well as for its evaluation regarding fairness, a warping approach must be found to create an approximation of the FiND world. 
This can be considered to be a pre-processing approach to fairML \shortcite<for more details on categorizing fairML approaches, see>{caton_fairness_2023}:
First, a warping function is learned by a suitable warping method and applied to the entire data. Then, a predictive model is tuned, trained, and evaluated on the warped data.
Finally, at prediction time, the feature vector of a new observation is warped by the same warping function and the predictive model is applied to that warped feature vector.

In the following, we use rank-preserving interventional distributions (RPID) to identify the FiND world and a residual-based warping method, both proposed by \shortciteA{bothmann_causal_2023} as warping method. We sketch the idea in the following and refer to the original source for details. This is just one example of a possible warping method \shortcite<another option could be to adapt quantile regression forests as proposed by>{plecko_fair_2020}, and it is left to future research to develop and compare more warping methods which can be embedded into our framework. 

\begin{figure}[ht]
     \centering
         \includegraphics[width=\textwidth]{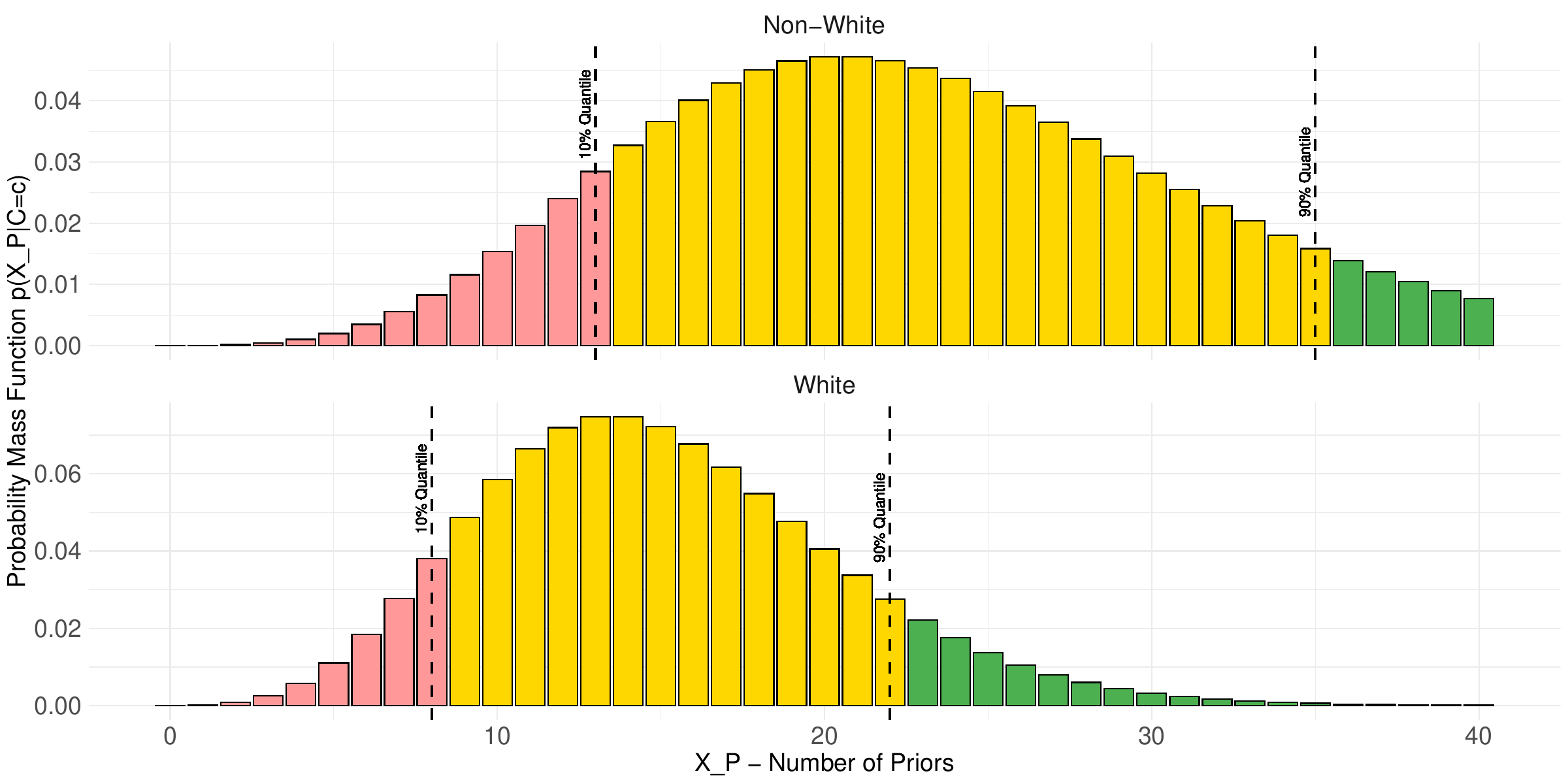}
         \caption{Illustration of RPID: Quantiles (here: $10\%$ and $90\%$) of the protected group (here: Non-White, top plot), are warped to their counterparts in the non-protected group (here: White, bottom plot). Fictitious distributions for illustration purposes.}
         \label{fig:rpid_viz}
\end{figure}

The core idea of RPID is to make the variables neutral with respect to the PA while maintaining any individual merits, sequentially, for all descendants of the PA. For example, a Non-White person, who is at the $90\%$-quantile of the confounder-specific distribution of $X_P$ in the Non-White subgroup will be “warped” to the $90\%$-quantile of the confounder-specific distribution of $X_P$ in the White subgroup, see also Figure \ref{fig:rpid_viz}.

RPID defines interventions on the SCM that lead to a FiND world that respects both the requirements on the PAs and individual merits. The estimands are defined on the joint counterfactual post-intervention distribution, e.g., the counterfactual $X_P$-distribution. Whether the desired estimand can be identified, i.e., expressed as a function of the observed data, requires considerations on, e.g., confounders of the mediator-outcome relationship. 
The concrete estimation method is based on a particular g-formula factorization to estimate the distributions of the FiND world. 
A model-based residual approach is used for estimating FiND world distributions and for warping real-world data: Comparably to \shortciteA{bothmann_causal_2023}, we assume an additive, homoscedastic error term in all models of mediators and target variable, given parents of these variables in the DAG, where we consider separate models per subgroup of the PA. 

This ultimately leads to a warping function from the real world to the warped world. Model training, validation, and prediction for new observations take place in the warped world.
This means that we 
focus on estimating a well-performing model, and try to achieve individual well-calibration -- but this time in the warped world. 
There is no longer a need for special evaluation measures for predictive performance in order to assess (un)fairness. The complexity is shifted to estimating the SCMs and the individual conditional quantiles. 
Note that it is of paramount importance for a successful practical use case to implement a close collaboration between subject-matter experts and machine learning experts, since expert knowledge will have to go hand-in-hand with empirical findings  \shortcite<as also pointed out by others before, e.g.,>{fleisher_algorithmic_2023,kasirzadeh_use_2021,kong_are_2022,loi_is_2022} in order to derive realistic causal models.
The practical usability of this approach depends on the extent to which the causal relationships in the real world can be estimated as well as on the success of removing certain causal relationships without modifying others.

\subsection{Experiments}
\label{sec:experiments}

As an example, we show experiments on the COMPAS data \shortcite{angwin_machine_2016}.\footnote{All experiments are fully reproducible, see \url{https://github.com/slds-lmu/paper_2024_wif}.} These experiments illustrate how our framework could be applied in practice and show how the warping approach affects the predictions for the different subgroups regarding the PA. However, this must not be misunderstood as a fully applied use case. As Section \ref{sec:appl-usecase} in the Appendix demonstrates, several normative questions must be answered carefully, followed by cautious modeling of the data. We use the DAGs of Figure \ref{fig:causal_graph}, noting again that this is just an illustration, and the true DAG will be more complex, contain more features, and must be developed together with domain experts.

We apply the residual-based warping approach outlined above: We sequentially warp $X_P$, $X_D$, and $Y$ of the Non-White individuals, using Poisson models for $X_P|A=a, \Cv$, and logit models for $X_D|A=a, \Cv$ and $Y|A=a, \Cv, X_D, X_P$, with $a\in \{\textit{Non-White}, \textit{White}\}$ (see Figure \ref{fig:causal_graph} for variable descriptions). 
In a nutshell, this works as follows, at the example of $X_P$: We train two Poisson models $\fh_n(\Cv)$ and $\fh_w(\Cv)$, for the Non-White and White subgroups, respectively. 
Then, we warp Non-White individual's $i$ value $\xpi$ in three steps: (i) compute its residual with respect to the Non-White model, i.e., $\ri = \xpi - \fh_n(\cvi)$; (ii) compute its relative position among all Non-White residuals for this model, e.g., 95\% quantile, and find the corresponding quantile $\qwi$ of White residuals for the White model $\fh_w(\cdot)$; 
(iii) warp $\xpi$ to the sum of this transformed residual and the White model prediction, i.e., $\xhpi = \fh_w(\cvi) + \qwi$. This is done sequentially for $X_D$ and $Y$ as well, where the already warped values for $X_D$ and $X_P$ ($\xhdi$ and $\xhpi$) are plugged into the prediction model of $Y$ in step (iii).\footnote{Technical note: While we use the exact warped features in this sequential algorithm, we show rounded versions in Table \ref{tab:warping} for better readability.}
As result, $i$'s features are warped from $\xi$ to $\xti$ and its target from $\yi$ to $\yti$.\footnote{At prediction time, where target $\yi$ is not available, we only warp the feature vector $\xi$.}

White individuals' features and target are not warped, i.e., $\xti = \xi$ and $\yti = \yi$, but the predicted recidivism rates also change from real-world rates $\pihi$ to warped-world rates $\pithi$ due to differing training data of Non-White individuals. As prediction models for the recidivism rate, we train logit models on real-world and warped data, respectively. We train all models (warping and recidivism rates) on an $80\%$ random subsample of the data and test on the remaining observations. Table \ref{tab:warping} shows the individuals of the test data where the predicted recidivism rate changed most between real world and warped world (column \textit{diff}). We can observe that (a) this is associated with a decrease from $X_P$ to $\tilde{X}_P$ for Non-White individuals who had the strongest change, and also that (b) the predicted recidivism probability is lower in the warped world for White individuals with a high value of $X_P$ (values range between 0 and 38 in the data). This is due to a lower coefficient in the logit model related to this feature, which is $0.15$ (se $0.0076$) in the real world and $0.05$ (se $0.0075$) in the warped world.

\begin{table}%
  \centering
  \subfloat[][]{    \begin{tabular}{rrrrrrrrr}
    \toprule
    \textit{gender} & \textit{age} & $X_P$ & $X_D$ & $\tilde{X}_P$ & $\tilde{X}_D$ & $\pihi$ & $\pithi$ & \textit{diff}\\
    \midrule
Male & 63 & 23 & 0 & 15 & 0 & 0.83 & 0.30 & -0.53 \\ 
Male & 53 & 26 & 0 & 18 & 0 & 0.93 & 0.41 & -0.52 \\ 
Male & 56 & 22 & 0 & 15 & 0 & 0.85 & 0.35 & -0.50 \\ 
   \dots & \dots & \dots & \dots & \dots &\dots &\dots & \dots & \dots \\ 
Male & 54 & 0 & 1 & 0 & 1 & 0.15 & 0.22 & 0.07 \\ 
Male & 55 & 0 & 1 & 0 & 1 & 0.15 & 0.22 & 0.07 \\ 
Male & 56 & 0 & 1 & 0 & 1 & 0.14 & 0.21 & 0.07 \\ 
    \bottomrule
    \end{tabular}}%
  \qquad
  \subfloat[][]{    \begin{tabular}{rrrrrrr}
    \toprule
     \textit{gender} & \textit{age} & $X_P$ & $X_D$ &  $\pihi$ & $\pithi$ & \textit{diff}\\
    \midrule
Female & 50 & 30 & 1 & 0.94 & 0.53 & -0.41 \\ 
Female & 47 & 28 & 0 & 0.94 & 0.54 & -0.40 \\ 
Male & 55 & 33 & 0 & 0.97 & 0.60 & -0.37 \\ 
  \dots & \dots &  \dots & \dots & \dots & \dots & \dots \\ 
Male & 52 & 0 & 1 & 0.17 & 0.26 & 0.09 \\ 
Male & 52 & 0 & 1 & 0.17 & 0.26 & 0.09 \\ 
Male & 53 & 0 & 1 & 0.16 & 0.26 & 0.10 \\ 
    \bottomrule
    \end{tabular}}
  \caption{Strongest warped individuals for (a) Non-White and (b) White individuals in test data, where $\textit{diff}= 
  \pithi - \pihi$. Individuals above the dots 
  are negatively discriminated in the real world, while individuals below the dots 
  are positively discriminated in the real world. Overall, 
  negative discrimination is higher for Non-White individuals and positive discrimination is higher for White individuals.}%
  \label{tab:warping}
\end{table}

\begin{figure}[th]
    \centering
    \begin{subfigure}[b]{0.49\textwidth}
         \includegraphics[width=\textwidth]{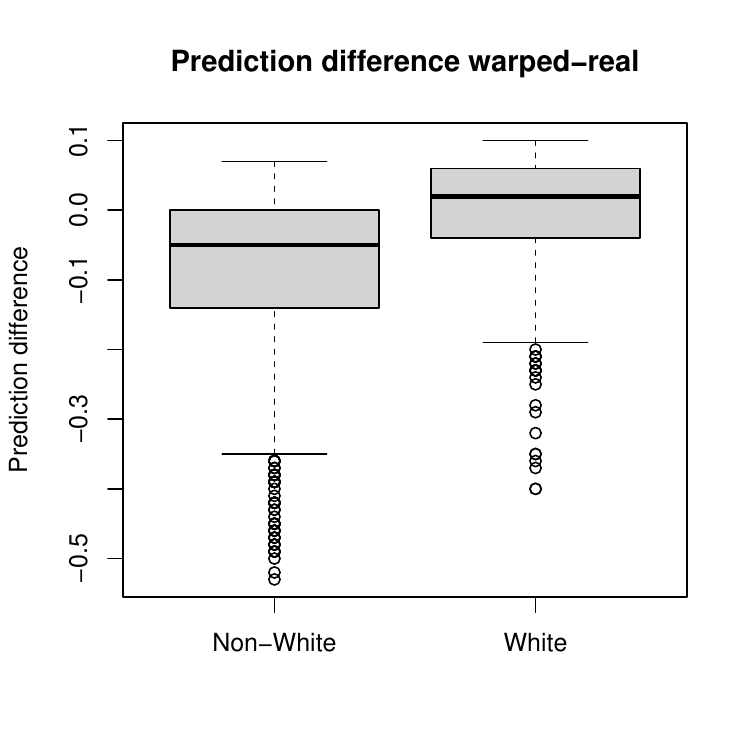}
         \caption{}
         \label{fig:compas_warped_preds}
     \end{subfigure}
     \begin{subfigure}[b]{0.49\textwidth}
         \includegraphics[width=\textwidth]{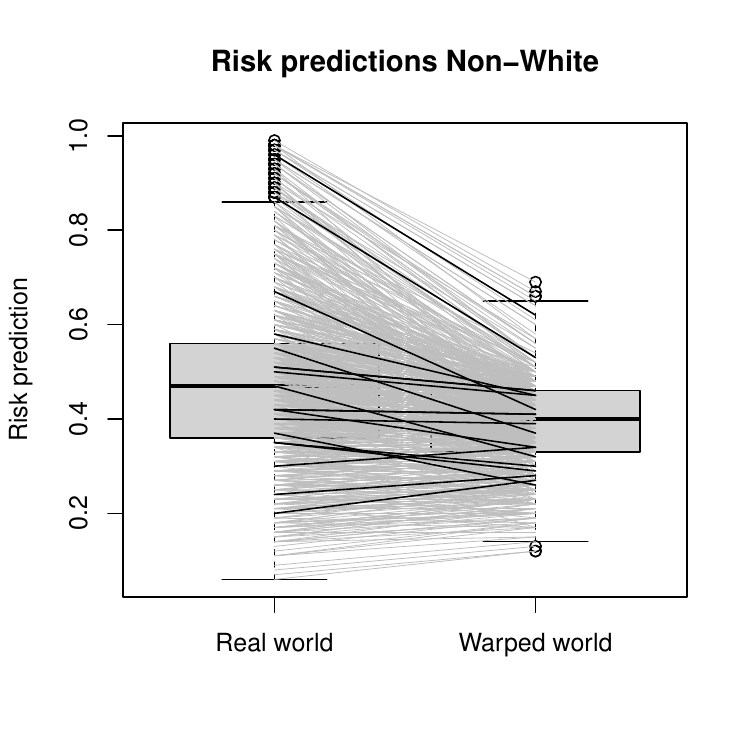}
         \caption{}
         \label{fig:compas_warped_preds_non-white}
     \end{subfigure}
     \caption{Recidivism predictions in the warped world and the real world. (a) shows differences in predictions for both groups and (b) shows predictions in both worlds for the Non-White group, where lines connect the same individuals (a random subsample has black lines for better readability).}
     \label{fig:compas_warped}
\end{figure}

Figure \ref{fig:compas_warped_preds} visualizes the prediction differences $\pithi - \pihi$ between the worlds for both PA groups. While White prediction differences vary around 0 (mean: $-0.0047$, p-value of t-test against 0: $0.254$), Non-White prediction differences show a trend towards lower values (mean: $-0.0918$, p-value of t-test: $<10^{-91}$), which means that -- on average -- their predicted recidivism rates are $9.2\%$ points lower in the warped world. While $76\%$ of Non-White individuals receive lower predictions in the warped world ($41\%$ for White), $24\%$ receive higher predictions ($59\%$ for White). Figure \ref{fig:compas_warped_preds_non-white} visualizes the predictions in both worlds for the Non-White subgroup. We see the overall negative trend, where the ranking within the group changes between the two worlds. This indicates differing individual strengths of discrimination in the real world and illustrates the individual perspective of our proposal. 

\begin{figure}[th]
    \centering
    \begin{subfigure}[b]{0.49\textwidth}
         \includegraphics[width=\textwidth]{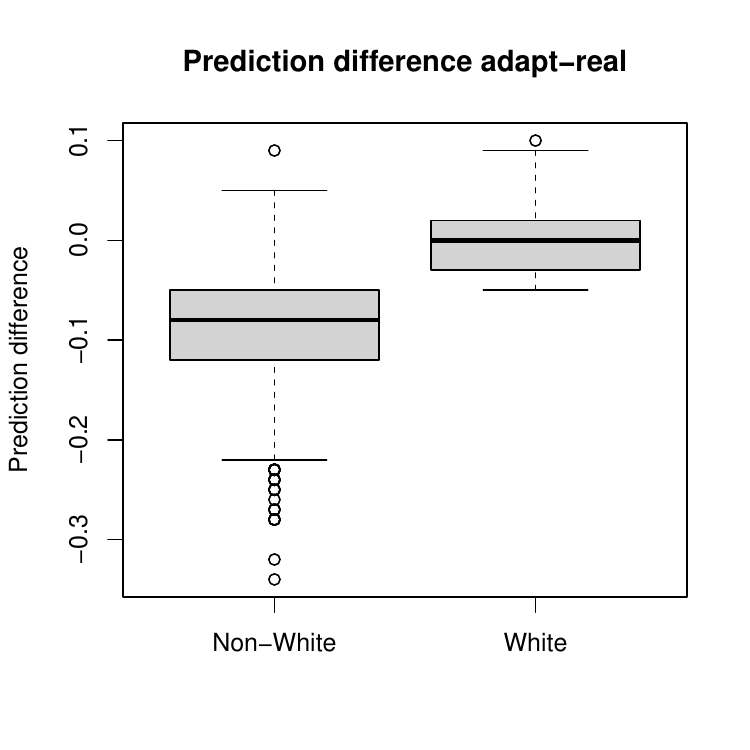}
         \caption{}
         \label{fig:compas_adapt_preds}
     \end{subfigure}
     \begin{subfigure}[b]{0.49\textwidth}
         \includegraphics[width=\textwidth]{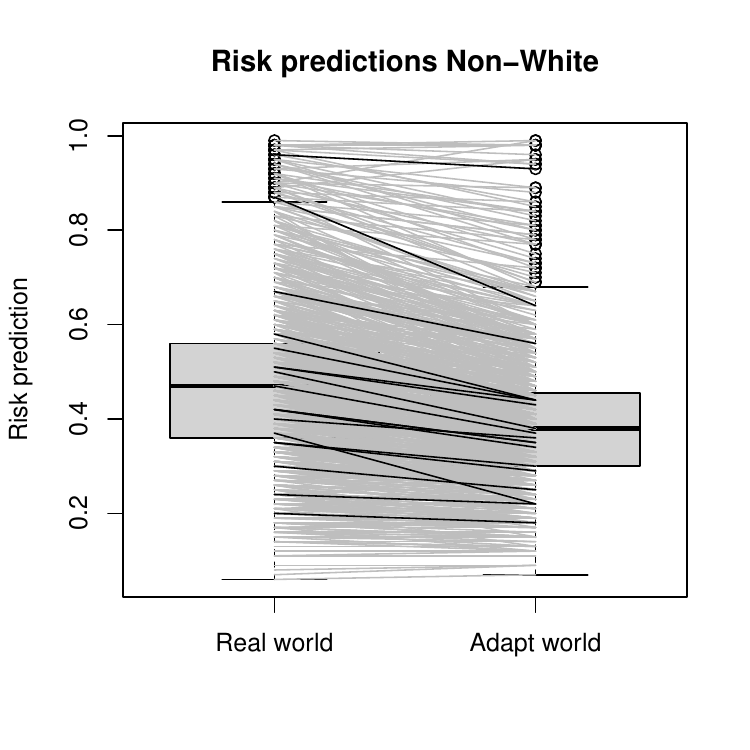}
         \caption{}
         \label{fig:compas_adapt_preds_non-white}
     \end{subfigure}
     \caption{Pendant to Figure \ref{fig:compas_warped}, showing recidivism predictions in the adapt world and the real world. (a) shows differences in predictions for both groups and (b) shows predictions in both worlds for the Non-White group, where lines connect the same individuals (a random subsample has black lines for better readability).}
\end{figure}

To illustrate the impact that the concrete choice of warping method can have on the results, we also applied ``fairadapt'' \shortcite{plecko_fair_2020} via the R package \texttt{fairadapt} \shortcite{plecko_r_package}:
The goal is to compute \textit{fair twins} via a \textit{do}-intervention on the PA, aiming to minimize ``the distortion in the data coming from the projection'', using quantile regression forests \shortcite{meinshausen_quantile_2006} as estimation method. 
Figure \ref{fig:compas_adapt_preds} shows the prediction differences between the real world and the ``adapt'' world for both PA groups. Similar to above, the prediction difference for the Non-White subgroups shifted toward negative values, with the values for the White subgroup varying around 0. Figure \ref{fig:compas_adapt_preds_non-white} shows the predictions in the real world and the adapt world for the Non-White subgroup. We see an overall negative trend as above, where the variability seems to be higher in the adapt world than in the warped world.


In an applied use case, we cannot prove or disprove that the warping worked and that the warped world is a good approximation of the FiND world. Similarly, we cannot judge whether the residual-based warping approach or fairadapt is superior. To make such a statement, we would need access to the FiND world. To properly investigate whether a warping method works in this sense, a thorough simulation study is necessary. Such a simulation study should also consider investigating other warping methods and comparing their behavior. Answering the question of how to optimally design a warping method to approximate the FiND world is important future work and beyond the scope of this paper. 
Nevertheless, the experiments presented here illustrate the type of results that can be expected from applying our framework in practice. Furthermore, the fact that the general direction of the results of the two warping approaches analyzed here are similar is an indication that both capture real-world discrimination and can potentially serve as bias-transforming methods.

\section{Conclusion and Discussion}
\label{sec:concl}
Despite a rapidly growing body of literature 
in fairML, a unified theoretical foundation was still lacking. By drawing on basic philosophical ideas, such a foundation to support the fairness debate was derived. 
This paper has identified the fundamental axioms that underpin the fairness debate to date, albeit without being made explicit, and has worked out their relationship to each other.
It was necessary to separate the basic structure of fairness from normative specifications for specific scenarios -- i.e., from material aspects.
Based on this, the normative stipulations could be precisely identified, which must take place outside ML.
In the case of PA-neutrality, 
three normative questions arise for the task at hand:
\begin{enumerate}
\item How is the task-specific merit $\wi$, i.e., the measure of task-specific equality of individuals, defined (e.g., by identifying it with an (un-)observable feature or with a function of several such features)? 
\item Which attributes are defined as PAs (if there are any)?
\item Shall the treatment function $s(\wi)$ be modified, and if so, how?  
\end{enumerate}

ML models provide estimates of the task-specific merit and are not unfair per se, 
but can induce unfairness.
Regardless of the presence of PAs, fairness problems can arise with ML, e.g., if the model is not individually well-calibrated. Thus, we do not see a tradeoff between fairness and predictive performance; rather, predictive performance is essential for fairness.
Since individual well-calibration cannot be checked empirically, approaches that consider all computationally identifiable subgroups should be pursued.
Once PAs are defined we move to 
the FiND world -- where causal effects of the PAs 
have been removed -- and the task of ML models is to predict the such modified counterparts of the real-world quantities in order to estimate the task-specific merit of individuals.
The concept of fairness is intrinsically related to causal considerations. 
However, 
causal questions are generally hard to answer with observational data \shortcite{kilbertus_avoiding_2017,kusner_counterfactual_2017,pearl_causal_2016}; for fairML, this means that causal approaches must come into focus even more. 

\paragraph{Limitations and Future Research} 
While we present first algorithms and a workflow for applying our framework in practice, working out detailed warping methods and concrete techniques to measure the validity of the warped data is left to future research.
This includes handling the inherent errors occurring by estimating the FiND world, such as the uncertainty regarding the true real-world DAG.

This paper considers fairness questions of ADM systems, where an ML model is used to predict some characteristic of interest in the subject matter concerned. Other applications of ML models -- such as diagnostic usage as advocated in, e.g., \shortciteA{barabas_interventions_2018} -- are out of scope. 
Considering PAs, we focus on how to reflect \myemph{PA-neutrality} in the ADM process. For future studies, we posit that a worthwhile direction is to tackle the incorporation of \myemph{PA-focus} as well (which, again, fits into the same basic philosophical concept of normative, task-specific equality but demands different technical solutions).

As mentioned above, the focus of this work is on the theoretical problem of fairML, pointing out weaknesses in current proposals of fairML and showing a direction to overcome these. 
We did not want to hide these fundamental challenges by presenting a fully applied use case, hence pretending that we would have the perfect solution for all challenges. Therefore, we limited ourselves to some illustrative examples that we considered helpful to understand our framework. 
The investigation of proper solutions for these challenges is important for future research in this field, and we hope that this work inspires other researchers to pursue this path, eventually leading to improved methods to deal with fairness issues in ML and ADM.

\newpage
\section*{Ethical Statements}


\paragraph{Ethical Considerations Statement}

Our work has a rather theoretical, philosophical perspective, and we do not propose a concrete application that could harm individuals. One goal of this paper is to disentangle normative choices and ML-related methodology to provide greater linguistic clarity when discussing fairness in ML and ADM systems. While conducting this work, it was important for us to be as specific as possible about this distinction between normative and methodological questions. We emphasize that the answers to these normative questions (as summarized in Section \ref{sec:concl}) must be the product of careful ethical considerations and, in many cases, the result of a broader societal discussion.

\paragraph{Researcher Positionality Statement}

Part of the author group is educated and based in the fields of statistics and computer science. While working in methodological research of ML, the authors appreciate philosophical questions surrounding the analysis of data and are skeptical of technical proposals that seem not to deeply interrogate \textit{why} they are proposing these solutions. The other part of the author group is educated and based in the fields of law and philosophy. They work on the very philosophical foundations of law and, thereby, also engage closely with ethical questions.
The interdisciplinary nature of our team contributes to a more comprehensive understanding of fairness in ML.

\paragraph{Adverse Impact Statement}

As for other technical solutions, blindly trusting techniques to overcome fairness problems can have negative societal impacts. However, we hope that by explicitly tasking non-ML users with answering three broadly understandable normative questions and by linking the ML model evaluation to the respective answers, we contribute a step in the direction of bringing societal needs and technical solutions closer together. Nevertheless, if the answers establish undesirable, e.g., discriminatory norms, then there is no technical possibility to remedy this. Therefore, it is of paramount importance that the normative stipulations -- especially in critical applications -- are transparent and result in a broad societal discussion.



\appendix

\section{Workflow for Applied Use Case}
\label{sec:appl-usecase}

\begin{figure}[th]
     \centering
         \includegraphics[width=\textwidth, trim=10 130 80 40, clip]{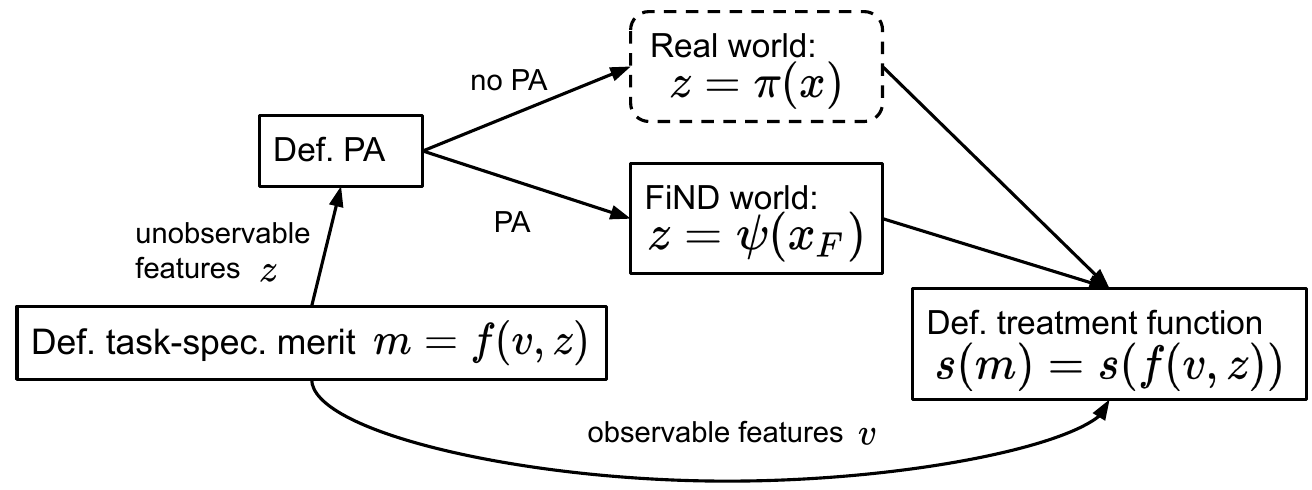}
         \caption{Workflow for an applied use case. Solid rectangles refer to normative decisions.}
         \label{fig:workflow}
\end{figure}

Figure \ref{fig:workflow} shows a workflow of how to use our framework in an applied ADM use case. Three normative questions arise, although it is important to note that these can never be answered empirically:

\begin{enumerate}
\item How is the task-specific merit $\wi$, i.e., the measure of task-specific equality of individuals, defined? 
\item Which attributes are defined as PAs (if there are any)?
\item Shall the treatment function $s(\wi)$ be modified, and if so, how?
\end{enumerate}

\paragraph{(1) Task-specific merit:} For the task at hand, the normative decision must be made of which features $\w$ are used to define individuals as equal. In other words, two individuals $i$ and $j$ are considered equal \textit{for this specific task} if and only if $\wi=\wj$ and, hence, have the same task-specific merit. The task-specific merit may consist of (a function of) observable $\vi$ and unobservable $\zi$ features, $\wi = f(\vi, \zi)$. In the COMPAS example, $\vi$ might be the type of crime and $\zi$ might be the two-year recidivism rate. In an example where individuals apply for credit in a bank, $\vi$ might be the amount of credit and $\zi$ the probability of default. In a tax example, there might be no unobservable variables because all features for the tax declaration are observable at decision time.

\paragraph{(2) Protected attributes:} Next, the normative decision must be made if there are PAs -- and if there are any PAs, to which feature they correspond. 
If applicable, the decision can be made to just remove some path-specific effects starting in the PA and to keep others in the FiND world (aiming for lawful indirect discrimination).
Together, this defines the FiND world.
These are delicate questions and surely the result of a broader, societal discussion; the presence of PAs means that for some groups, the unobservable features are not used as-is, but they will be transformed into their FiND world counterpart.  This might potentially change the (numerical) task-specific merit of some or all individuals and have a large impact on how people are ranked overall (as exemplified in the experiments in Section \ref{sec:experiments}). As a result, without PAs, the unobservable features will be identified with the estimand in the real world, i.e., $\zi = \pixii$, whereas in the presence of PAs, the unobservable features will be identified with the estimand in the FiND world, i.e., $\zi = \psi(\xfindi)$.

\paragraph{(3) Treatment function:} Finally, the treatment function $s(\wi) = s(f(\vi, \zi))$ must be defined normatively. Again, this results from a societal or expert discussion regarding the use case at hand. In the COMPAS example, a binary decision on concrete treatment might be made using a threshold on the recidivism probability, where the threshold varies between different types of crime. In the credit example, a binary decision on granting credit might be made using a threshold on the probability of default, where the threshold varies between different amounts of credit. In the tax rate example, the decision only relies on observable features and is subject to regular political discussion.\\

Once these questions are answered, algorithms as outlined in Section \ref{sec:implementation} can be used to approximate the FiND world via warping. Then, warped-world data are used to train ML models and to predict from these trained ML models for new, unseen individuals -- after warping their feature vector.\\

In summary, the task of subject matter researchers is to define the PAs, the DAG in the real world, and other variables that are part of the task-specific merit (outside of the DAG), where the tasks of ML experts are to find good warping methods to approximate the such defined FiND world and to train well-performing prediction models in the warped world.\\

\paragraph{Remark on PAs and observable features} Figure \ref{fig:workflow} shows that observable features are used as-is, no matter if there are PAs. However, the normative result of a discussion on this could be that not only the unobservable but also the observable features may be subject to historic unfairness (e.g., income with the PA gender) and that these features should also be transferred to the FiND world (e.g., asking ``what would the income be in a world where gender has no causal effect on income''). This is a straightforward extension of our framework: Instead of using the observable feature $v$ directly, it will be replaced by its FiND world counterpart $v_F$, which then must be estimated, e.g., by an ML model, yielding $\hat{{v}}$.

\bibliography{mybib, mybib-k, mybib-manuell}

\begin{thebibliography}{}

\bibitem[\protect\BCAY{Agarwal, Beygelzimer, Dudik, Langford,\ \BBA\
  Wallach}{Agarwal et~al.}{2018}]{agarwal_reductions_2018}
Agarwal, A., Beygelzimer, A., Dudik, M., Langford, J., \BBA\ Wallach, H.
  \BBOP2018\BBCP.
\newblock \BBOQ A {Reductions} {Approach} to {Fair} {Classification}\BBCQ\
\newblock In Dy, J.\BBACOMMA\  \BBA\ Krause, A.\BEDS, {\Bem Proceedings of the
  35th {International} {Conference} on {Machine} {Learning}}, \BPGS\ 60--69.
  PMLR.

\bibitem[\protect\BCAY{Angwin, Larson, Mattu,\ \BBA\ Kirchner}{Angwin
  et~al.}{2016}]{angwin_machine_2016}
Angwin, J., Larson, J., Mattu, S., \BBA\ Kirchner, L. \BBOP2016\BBCP.
\newblock \BBOQ Machine {Bias}: {There}’s {Software} {Used} {Across} the
  {Country} to {Predict} {Future} {Criminals}, and {It}’s {Biased} {Against}
  {Blacks}\BBCQ\
\newblock \textit{ProPublica}.

\bibitem[\protect\BCAY{Aristotle}{Aristotle}{2009}]{aristotle_nicomachean_2009}
Aristotle \BBOP2009\BBCP.
\newblock {\Bem The {Nicomachean} ethics (book {V})}.
\newblock Oxford {World}'s {Classics}. Oxford University Press.

\bibitem[\protect\BCAY{Baggio, Corsini, Floreani, Giannini,\ \BBA\
  Zagonel}{Baggio et~al.}{2013}]{baggio_gender_2013}
Baggio, G., Corsini, A., Floreani, A., Giannini, S., \BBA\ Zagonel, V.
  \BBOP2013\BBCP.
\newblock \BBOQ Gender {Medicine}: {A} {Task} for the {Third}
  {Millennium}\BBCQ\
\newblock {\Bem Clinical Chemistry and Laboratory Medicine (CCLM)}, {\Bem
  51\/}(4), 713--727.

\bibitem[\protect\BCAY{Barabas, Virza, Dinakar, Ito,\ \BBA\ Zittrain}{Barabas
  et~al.}{2018}]{barabas_interventions_2018}
Barabas, C., Virza, M., Dinakar, K., Ito, J., \BBA\ Zittrain, J.
  \BBOP2018\BBCP.
\newblock \BBOQ Interventions over {Predictions}: {Reframing} the {Ethical}
  {Debate} for {Actuarial} {Risk} {Assessment}\BBCQ\
\newblock In {\Bem Proceedings of the 1st {Conference} on {Fairness},
  {Accountability} and {Transparency}}, \BPGS\ 62--76. PMLR.

\bibitem[\protect\BCAY{Barocas, Hardt,\ \BBA\ Narayanan}{Barocas
  et~al.}{2019}]{barocas_fairness_2019}
Barocas, S., Hardt, M., \BBA\ Narayanan, A. \BBOP2019\BBCP.
\newblock {\Bem Fairness and {Machine} {Learning}}.
\newblock MIT Press.

\bibitem[\protect\BCAY{Bechavod, Jung,\ \BBA\ Wu}{Bechavod
  et~al.}{2020}]{bechavod_metric-free_2020}
Bechavod, Y., Jung, C., \BBA\ Wu, S.~Z. \BBOP2020\BBCP.
\newblock \BBOQ Metric-{Free} {Individual} {Fairness} in {Online}
  {Learning}\BBCQ\
\newblock In Larochelle, H., Ranzato, M., Hadsell, R., Balcan, M.~F., \BBA\
  Lin, H.\BEDS, {\Bem Advances in {Neural} {Information} {Processing}
  {Systems}}, \lowercase{\BVOL}~33, \BPGS\ 11214--11225. Curran Associates,
  Inc.

\bibitem[\protect\BCAY{Beck, Grunwald, Jacob,\ \BBA\ Matzner}{Beck
  et~al.}{2019}]{beck_kunstliche_2019}
Beck, S., Grunwald, A., Jacob, K., \BBA\ Matzner, T. \BBOP2019\BBCP.
\newblock {\Bem Künstliche {Intelligenz} und {Diskriminierung}}.
\newblock Lernende Systeme, München.

\bibitem[\protect\BCAY{Berk, Heidari, Jabbari, Kearns,\ \BBA\ Roth}{Berk
  et~al.}{2021}]{berk_fairness_2021}
Berk, R., Heidari, H., Jabbari, S., Kearns, M., \BBA\ Roth, A. \BBOP2021\BBCP.
\newblock \BBOQ Fairness in {Criminal} {Justice} {Risk} {Assessments}: {The}
  {State} of the {Art}\BBCQ\
\newblock {\Bem Sociological Methods \& Research}, {\Bem 50\/}(1), 3--44.
\newblock SAGE Publications Inc.

\bibitem[\protect\BCAY{Binns}{Binns}{2018}]{binns_fairness_2018}
Binns, R. \BBOP2018\BBCP.
\newblock \BBOQ Fairness in {Machine} {Learning}: {Lessons} from {Political}
  {Philosophy}\BBCQ\
\newblock In Friedler, S.~A.\BBACOMMA\  \BBA\ Wilson, C.\BEDS, {\Bem
  Proceedings of the 1st {Conference} on {Fairness}, {Accountability} and
  {Transparency}}, \BPGS\ 149--159. PMLR.

\bibitem[\protect\BCAY{Black, Elzayn, Chouldechova, Goldin,\ \BBA\ Ho}{Black
  et~al.}{2022}]{black_algorithmic_2022}
Black, E., Elzayn, H., Chouldechova, A., Goldin, J., \BBA\ Ho, D.
  \BBOP2022\BBCP.
\newblock \BBOQ Algorithmic {Fairness} and {Vertical} {Equity}: {Income}
  {Fairness} with {IRS} {Tax} {Audit} {Models}\BBCQ\
\newblock In {\Bem 2022 {ACM} {Conference} on {Fairness}, {Accountability}, and
  {Transparency}}, \BPGS\ 1479--1503, Seoul Republic of Korea. ACM.

\bibitem[\protect\BCAY{Bothmann, Dandl,\ \BBA\ Schomaker}{Bothmann
  et~al.}{2023}]{bothmann_causal_2023}
Bothmann, L., Dandl, S., \BBA\ Schomaker, M. \BBOP2023\BBCP.
\newblock \BBOQ Causal {Fair} {Machine} {Learning} via {Rank}-{Preserving}
  {Interventional} {Distributions}\BBCQ\
\newblock In {\Bem Proceedings of the 1st {Workshop} on {Fairness} and {Bias}
  in {AI} co-located with 26th {European} {Conference} on {Artificial}
  {Intelligence} ({ECAI} 2023)}. CEUR Workshop Proceedings.

\bibitem[\protect\BCAY{Calsamiglia}{Calsamiglia}{2005}]{calsamiglia_decentralizing_2005}
Calsamiglia, C. \BBOP2005\BBCP.
\newblock {\Bem Decentralizing {Equality} of {Opportunity} and {Issues}
  {Concerning} the {Equality} of {Educational} {Opportunity}}.
\newblock Yale University.

\bibitem[\protect\BCAY{{Cambridge Dictionary}}{{Cambridge
  Dictionary}}{2022}]{cambridge_dictionary_fairness}
{Cambridge Dictionary} \BBOP2022\BBCP.
\newblock {\Bem Fairness}.
\newblock Cambridge University Press \& Assessment.

\bibitem[\protect\BCAY{Caton\ \BBA\ Haas}{Caton\ \BBA\
  Haas}{2024}]{caton_fairness_2023}
Caton, S.\BBACOMMA\  \BBA\ Haas, C. \BBOP2024\BBCP.
\newblock \BBOQ Fairness in {Machine} {Learning}: {A} {Survey}\BBCQ\
\newblock \textit{ACM Computing Surveys}, 3616865.

\bibitem[\protect\BCAY{Chiappa}{Chiappa}{2019}]{chiappa_path-specific_2019}
Chiappa, S. \BBOP2019\BBCP.
\newblock \BBOQ Path-{Specific} {Counterfactual} {Fairness}\BBCQ\
\newblock {\Bem Proceedings of the AAAI Conference on Artificial Intelligence},
  {\Bem 33\/}(01), 7801--7808.

\bibitem[\protect\BCAY{Chiappa\ \BBA\ Isaac}{Chiappa\ \BBA\
  Isaac}{2019}]{chiappa_causal_2019}
Chiappa, S.\BBACOMMA\  \BBA\ Isaac, W.~S. \BBOP2019\BBCP.
\newblock \BBOQ A {Causal} {Bayesian} {Networks} {Viewpoint} on
  {Fairness}\BBCQ\
\newblock In {\Bem Privacy and {Identity} {Management}. {Fairness},
  {Accountability}, and {Transparency} in the {Age} of {Big} {Data}},
  \lowercase{\BVOL}\ 547, \BPGS\ 3--20. Springer International Publishing,
  Cham.
\newblock Series Title: IFIP Advances in Information and Communication
  Technology.

\bibitem[\protect\BCAY{Chikahara, Sakaue, Fujino,\ \BBA\ Kashima}{Chikahara
  et~al.}{2021}]{chikahara_learning_2021}
Chikahara, Y., Sakaue, S., Fujino, A., \BBA\ Kashima, H. \BBOP2021\BBCP.
\newblock \BBOQ Learning {Individually} {Fair} {Classifier} with
  {Path}-{Specific} {Causal}-{Effect} {Constraint}\BBCQ\
\newblock In {\Bem Proceedings of {The} 24th {International} {Conference} on
  {Artificial} {Intelligence} and {Statistics}}, \BPGS\ 145--153. PMLR.

\bibitem[\protect\BCAY{Chouldechova\ \BBA\ Roth}{Chouldechova\ \BBA\
  Roth}{2020}]{chouldechova_snapshot_2020}
Chouldechova, A.\BBACOMMA\  \BBA\ Roth, A. \BBOP2020\BBCP.
\newblock \BBOQ A snapshot of the frontiers of fairness in machine
  learning\BBCQ\
\newblock {\Bem Communications of the ACM}, {\Bem 63\/}(5), 82--89.

\bibitem[\protect\BCAY{Clarke}{Clarke}{2019}]{clarke_text_2019}
Clarke, Y.~D. \BBOP2019\BBCP.
\newblock \BBOQ Text - {H}.{R}.2231 - 116th {Congress} (2019-2020):
  {Algorithmic} {Accountability} {Act} of 2019\BBCQ\
\newblock Archive Location: 2019/2020.

\bibitem[\protect\BCAY{Coate\ \BBA\ Loury}{Coate\ \BBA\
  Loury}{1993}]{coate_will_1993}
Coate, S.\BBACOMMA\  \BBA\ Loury, G. \BBOP1993\BBCP.
\newblock \BBOQ Will {Affirmative}-{Action} {Policies} {Eliminate} {Negative}
  {Stereotypes}?\BBCQ\
\newblock {\Bem American Economic Review}, {\Bem 83\/}(5), 1220--40.

\bibitem[\protect\BCAY{Dator}{Dator}{2017}]{dator_chapter_2017}
Dator, J. \BBOP2017\BBCP.
\newblock \BBOQ Chapter 3. {What} {Is} {Fairness}?\BBCQ\
\newblock In Dator, J., Pratt, R.~C., \BBA\ Seo, Y.\BEDS, {\Bem Fairness,
  {Globalization}, and {Public} {Institutions}}, \BPGS\ 19--34. University of
  Hawaii Press, Honolulu.

\bibitem[\protect\BCAY{Dwork, Hardt, Pitassi, Reingold,\ \BBA\ Zemel}{Dwork
  et~al.}{2012}]{dwork_fairness_2012}
Dwork, C., Hardt, M., Pitassi, T., Reingold, O., \BBA\ Zemel, R.
  \BBOP2012\BBCP.
\newblock \BBOQ Fairness {Through} {Awareness}\BBCQ\
\newblock In {\Bem Proceedings of the 3rd {Innovations} in {Theoretical}
  {Computer} {Science} {Conference}}, \BPGS\ 214--226, New York, NY, USA.
  Association for Computing Machinery.

\bibitem[\protect\BCAY{EU}{EU}{2016}]{eu_eur-lex_2016}
EU \BBOP2016\BBCP.
\newblock \BBOQ {EUR}-{Lex} - {32016R0679} - {EN} - {EUR}-{Lex}\BBCQ.
\newblock Doc ID: 32016R0679 Doc Sector: 3 Doc Title: Regulation (EU) 2016/679
  of the European Parliament and of the Council of 27 April 2016 on the
  protection of natural persons with regard to the processing of personal data
  and on the free movement of such data, and repealing Directive 95/46/EC
  (General Data Protection Regulation).

\bibitem[\protect\BCAY{{Executive Office of the President}, Muñoz, Smith,\
  \BBA\ Patil}{{Executive Office of the President}
  et~al.}{2016}]{executive_office_2016}
{Executive Office of the President}, Muñoz, C., Smith, M., \BBA\ Patil, D.
  \BBOP2016\BBCP.
\newblock \BBOQ Big {Data}: {A} {Report} on {Algorithmic} {Systems},
  {Opportunity}, and {Civil} {Rights}\BBCQ.

\bibitem[\protect\BCAY{Fleisher}{Fleisher}{2023}]{fleisher_algorithmic_2023}
Fleisher, W. \BBOP2023\BBCP.
\newblock \BBOQ Algorithmic {Fairness} {Criteria} as {Evidence}\BBCQ\
\newblock \textit{SSRN}.

\bibitem[\protect\BCAY{Foster\ \BBA\ Vohra}{Foster\ \BBA\
  Vohra}{1992}]{foster_economic_1992}
Foster, D.~P.\BBACOMMA\  \BBA\ Vohra, R.~V. \BBOP1992\BBCP.
\newblock \BBOQ An {Economic} {Argument} for {Affirmative} {Action}\BBCQ\
\newblock {\Bem Rationality and Society}, {\Bem 4\/}(2), 176--188.

\bibitem[\protect\BCAY{Friedler, Scheidegger,\ \BBA\
  Venkatasubramanian}{Friedler et~al.}{2021}]{friedler_impossibility_2016}
Friedler, S.~A., Scheidegger, C., \BBA\ Venkatasubramanian, S. \BBOP2021\BBCP.
\newblock \BBOQ The (im)possibility of fairness: different value systems
  require different mechanisms for fair decision making\BBCQ\
\newblock {\Bem Commun. ACM}, {\Bem 64\/}(4), 136–143.

\bibitem[\protect\BCAY{Gajane\ \BBA\ Pechenizkiy}{Gajane\ \BBA\
  Pechenizkiy}{2017}]{gajane_formalizing_2017}
Gajane, P.\BBACOMMA\  \BBA\ Pechenizkiy, M. \BBOP2017\BBCP.
\newblock \BBOQ On {Formalizing} {Fairness} in {Prediction} {With} {Machine}
  {Learning}\BBCQ\
\newblock \textit{arXiv:1710.03184}.

\bibitem[\protect\BCAY{Grabowicz, Perello,\ \BBA\ Mishra}{Grabowicz
  et~al.}{2022}]{grabowicz_marrying_2022}
Grabowicz, P.~A., Perello, N., \BBA\ Mishra, A. \BBOP2022\BBCP.
\newblock \BBOQ Marrying {Fairness} and {Explainability} in {Supervised}
  {Learning}\BBCQ\
\newblock In {\Bem 2022 {ACM} {Conference} on {Fairness}, {Accountability}, and
  {Transparency}}, {FAccT} '22, \BPGS\ 1905--1916, New York, NY, USA.
  Association for Computing Machinery.

\bibitem[\protect\BCAY{Green}{Green}{2018}]{green_fair_2018}
Green, B. \BBOP2018\BBCP.
\newblock \BBOQ Fair {Risk} {Assessments}: {A} {Precarious} {Approach} for
  {Criminal} {Justice} {Reform}\BBCQ\
\newblock \textit{5th Workshop on Fairness, Accountability, and Transparency in
  Machine Learning}.

\bibitem[\protect\BCAY{Green}{Green}{2022}]{green_escaping_2022}
Green, B. \BBOP2022\BBCP.
\newblock \BBOQ Escaping the {Impossibility} of {Fairness}: {From} {Formal} to
  {Substantive} {Algorithmic} {Fairness}\BBCQ\
\newblock {\Bem Philosophy \& Technology}, {\Bem 35\/}(4), 90.

\bibitem[\protect\BCAY{Guion}{Guion}{1966}]{guion_employment_1966}
Guion, R.~M. \BBOP1966\BBCP.
\newblock \BBOQ Employment {Tests} and {Discriminatory} {Hiring}\BBCQ\
\newblock {\Bem Industrial Relations: A Journal of Economy and Society}, {\Bem
  5\/}(2), 20--37.
\newblock Publisher: Wiley Online Library.

\bibitem[\protect\BCAY{Hebert-Johnson, Kim, Reingold,\ \BBA\
  Rothblum}{Hebert-Johnson et~al.}{2018}]{hebert-johnson_multicalibration_2018}
Hebert-Johnson, U., Kim, M., Reingold, O., \BBA\ Rothblum, G. \BBOP2018\BBCP.
\newblock \BBOQ Multicalibration: {Calibration} for the
  ({Computationally}-{Identifiable}) {Masses}\BBCQ\
\newblock In Dy, J.\BBACOMMA\  \BBA\ Krause, A.\BEDS, {\Bem Proceedings of the
  35th {International} {Conference} on {Machine} {Learning}}, \BPGS\
  1939--1948. PMLR.

\bibitem[\protect\BCAY{Hedden}{Hedden}{2021}]{hedden_statistical_2021}
Hedden, B. \BBOP2021\BBCP.
\newblock \BBOQ On {Statistical} {Criteria} of {Algorithmic} {Fairness}\BBCQ\
\newblock {\Bem Philosophy \& Public Affairs}, {\Bem 49\/}(2), 209--231.

\bibitem[\protect\BCAY{Hutchinson\ \BBA\ Mitchell}{Hutchinson\ \BBA\
  Mitchell}{2019}]{hutchinson_50_2019}
Hutchinson, B.\BBACOMMA\  \BBA\ Mitchell, M. \BBOP2019\BBCP.
\newblock \BBOQ 50 {Years} of {Test} ({Un}){Fairness}: {Lessons} for {Machine}
  {Learning}\BBCQ\
\newblock In {\Bem Proceedings of the {Conference} on {Fairness},
  {Accountability}, and {Transparency}}, {FAT}* '19, \BPGS\ 49--58, New York,
  NY, USA. Association for Computing Machinery.

\bibitem[\protect\BCAY{Hütt\ \BBA\ Schubert}{Hütt\ \BBA\
  Schubert}{2020}]{mainzer_fairness_2020}
Hütt, M.-T.\BBACOMMA\  \BBA\ Schubert, C. \BBOP2020\BBCP.
\newblock \BBOQ Fairness von {KI}-{Algorithmen}\BBCQ\
\newblock In Mainzer, K.\BED, {\Bem Philosophisches {Handbuch} {Künstliche}
  {Intelligenz}}, \BPGS\ 1--22. Springer Fachmedien Wiesbaden, Wiesbaden.

\bibitem[\protect\BCAY{Kalev, Dobbin,\ \BBA\ Kelly}{Kalev
  et~al.}{2006}]{kalev_best_2006}
Kalev, A., Dobbin, F., \BBA\ Kelly, E. \BBOP2006\BBCP.
\newblock \BBOQ Best {Practices} or {Best} {Guesses}? {Assessing} the
  {Efficacy} of {Corporate} {Affirmative} {Action} and {Diversity}
  {Policies}\BBCQ\
\newblock {\Bem American Sociological Review}, {\Bem 71\/}(4), 589--617.

\bibitem[\protect\BCAY{Kasirzadeh\ \BBA\ Smart}{Kasirzadeh\ \BBA\
  Smart}{2021}]{kasirzadeh_use_2021}
Kasirzadeh, A.\BBACOMMA\  \BBA\ Smart, A. \BBOP2021\BBCP.
\newblock \BBOQ The {Use} and {Misuse} of {Counterfactuals} in {Ethical}
  {Machine} {Learning}\BBCQ\
\newblock In {\Bem Proceedings of the 2021 {ACM} {Conference} on {Fairness},
  {Accountability}, and {Transparency}}, \BPGS\ 228--236, Virtual Event Canada.
  ACM.

\bibitem[\protect\BCAY{Kasy\ \BBA\ Abebe}{Kasy\ \BBA\
  Abebe}{2021}]{kasy_fairness_2021}
Kasy, M.\BBACOMMA\  \BBA\ Abebe, R. \BBOP2021\BBCP.
\newblock \BBOQ Fairness, {Equality}, and {Power} in {Algorithmic}
  {Decision}-{Making}\BBCQ\
\newblock In {\Bem Proceedings of the 2021 {ACM} {Conference} on {Fairness},
  {Accountability}, and {Transparency}}, \BPGS\ 576--586, Virtual Event Canada.
  ACM.

\bibitem[\protect\BCAY{Kearns, Neel, Roth,\ \BBA\ Wu}{Kearns
  et~al.}{2018}]{kearns_preventing_2018}
Kearns, M., Neel, S., Roth, A., \BBA\ Wu, Z.~S. \BBOP2018\BBCP.
\newblock \BBOQ Preventing {Fairness} {Gerrymandering}: {Auditing} and
  {Learning} for {Subgroup} {Fairness}\BBCQ\
\newblock In Dy, J.\BBACOMMA\  \BBA\ Krause, A.\BEDS, {\Bem Proceedings of the
  35th {International} {Conference} on {Machine} {Learning}}, \BPGS\
  2564--2572. PMLR.

\bibitem[\protect\BCAY{Keith, Bell, Swanson,\ \BBA\ Williams}{Keith
  et~al.}{1985}]{keith_effects_1985}
Keith, S.~N., Bell, R.~M., Swanson, A.~G., \BBA\ Williams, A.~P.
  \BBOP1985\BBCP.
\newblock \BBOQ Effects of {Affirmative} {Action} in {Medical} {Schools}\BBCQ\
\newblock {\Bem New England Journal of Medicine}, {\Bem 313\/}(24), 1519--1525.

\bibitem[\protect\BCAY{Kilbertus, Rojas~Carulla, Parascandolo, Hardt, Janzing,\
  \BBA\ Schölkopf}{Kilbertus et~al.}{2017}]{kilbertus_avoiding_2017}
Kilbertus, N., Rojas~Carulla, M., Parascandolo, G., Hardt, M., Janzing, D.,
  \BBA\ Schölkopf, B. \BBOP2017\BBCP.
\newblock \BBOQ Avoiding {Discrimination} through {Causal} {Reasoning}\BBCQ\
\newblock In {\Bem Advances in {Neural} {Information} {Processing} {Systems}},
  \lowercase{\BVOL}~30. Curran Associates, Inc.

\bibitem[\protect\BCAY{Kim, Ghorbani,\ \BBA\ Zou}{Kim
  et~al.}{2019}]{kim_multiaccuracy_2019}
Kim, M.~P., Ghorbani, A., \BBA\ Zou, J. \BBOP2019\BBCP.
\newblock \BBOQ Multiaccuracy: {Black}-box {Post}-processing for {Fairness} in
  {Classification}\BBCQ\
\newblock In {\Bem Proceedings of the 2019 {AAAI}/{ACM} {Conference} on {AI},
  {Ethics}, and {Society}}, \BPGS\ 247--254, New York, NY, USA. Association for
  Computing Machinery.

\bibitem[\protect\BCAY{Kleinberg, Mullainathan,\ \BBA\ Raghavan}{Kleinberg
  et~al.}{2017}]{kleinberg_inherent_2017}
Kleinberg, J., Mullainathan, S., \BBA\ Raghavan, M. \BBOP2017\BBCP.
\newblock \BBOQ Inherent {Trade}-{Offs} in the {Fair} {Determination} of {Risk}
  {Scores}\BBCQ\
\newblock In Papadimitriou, C.~H.\BED, {\Bem 8th {Innovations} in {Theoretical}
  {Computer} {Science} {Conference} ({ITCS} 2017)}, \lowercase{\BVOL}~67 of
  {\Bem Leibniz {International} {Proceedings} in {Informatics} ({LIPIcs})},
  \BPGS\ 43:1--43:23, Dagstuhl, Germany. Schloss Dagstuhl–Leibniz-Zentrum
  fuer Informatik.

\bibitem[\protect\BCAY{Kong}{Kong}{2022}]{kong_are_2022}
Kong, Y. \BBOP2022\BBCP.
\newblock \BBOQ Are “{Intersectionally} {Fair}” {AI} {Algorithms} {Really}
  {Fair} to {Women} of {Color}? {A} {Philosophical} {Analysis}\BBCQ\
\newblock In {\Bem 2022 {ACM} {Conference} on {Fairness}, {Accountability}, and
  {Transparency}}, \BPGS\ 485--494, Seoul Republic of Korea. ACM.

\bibitem[\protect\BCAY{Kuppler, Kern, Bach,\ \BBA\ Kreuter}{Kuppler
  et~al.}{2021}]{kuppler_distributive_2021}
Kuppler, M., Kern, C., Bach, R.~L., \BBA\ Kreuter, F. \BBOP2021\BBCP.
\newblock \BBOQ Distributive {Justice} and {Fairness} {Metrics} in {Automated}
  {Decision}-making: {How} {Much} {Overlap} {Is} {There}?\BBCQ\
\newblock \textit{arXiv:2105.01441}.

\bibitem[\protect\BCAY{Kusner, Loftus, Russell,\ \BBA\ Silva}{Kusner
  et~al.}{2017}]{kusner_counterfactual_2017}
Kusner, M.~J., Loftus, J., Russell, C., \BBA\ Silva, R. \BBOP2017\BBCP.
\newblock \BBOQ Counterfactual {Fairness}\BBCQ\
\newblock In Guyon, I., Luxburg, U.~V., Bengio, S., Wallach, H., Fergus, R.,
  Vishwanathan, S., \BBA\ Garnett, R.\BEDS, {\Bem Advances in {Neural}
  {Information} {Processing} {Systems}}, \lowercase{\BVOL}~30. Curran
  Associates, Inc.

\bibitem[\protect\BCAY{Kusner\ \BBA\ Loftus}{Kusner\ \BBA\
  Loftus}{2020}]{kusner_long_2020}
Kusner, M.~J.\BBACOMMA\  \BBA\ Loftus, J.~R. \BBOP2020\BBCP.
\newblock \BBOQ The {Long} {Road} to {Fairer} {Algorithms}\BBCQ\
\newblock {\Bem Nature}, {\Bem 578\/}(7793), 34--36.

\bibitem[\protect\BCAY{Larson, Mattu, Kirchner,\ \BBA\ Angwin}{Larson
  et~al.}{2016}]{larson_how_2016}
Larson, J., Mattu, S., Kirchner, L., \BBA\ Angwin, J. \BBOP2016\BBCP.
\newblock \BBOQ How {We} {Analyzed} the {COMPAS} {Recidivism} {Algorithm}\BBCQ\
\newblock \textit{ProPublica}.

\bibitem[\protect\BCAY{Lee, Floridi,\ \BBA\ Singh}{Lee
  et~al.}{2020}]{lee_fairness_2020}
Lee, M., Floridi, L., \BBA\ Singh, J. \BBOP2020\BBCP.
\newblock \BBOQ From {Fairness} {Metrics} to {Key} {Ethics} {Indicators}
  ({KEIs}): {A} {Context}-{Aware} {Approach} to {Algorithmic} {Ethics} in an
  {Unequal} {Society}\BBCQ\
\newblock \textit{SSRN Electronic Journal}.

\bibitem[\protect\BCAY{Lee, Floridi,\ \BBA\ Singh}{Lee
  et~al.}{2021}]{lee_formalising_2021}
Lee, M. S.~A., Floridi, L., \BBA\ Singh, J. \BBOP2021\BBCP.
\newblock \BBOQ Formalising {Trade}-offs {Beyond} {Algorithmic} {Fairness}:
  {Lessons} from {Ethical} {Philosophy} and {Welfare} {Economics}\BBCQ\
\newblock {\Bem AI and Ethics}, {\Bem 1\/}(4), 529--544.

\bibitem[\protect\BCAY{Loi\ \BBA\ Heitz}{Loi\ \BBA\ Heitz}{2022}]{loi_is_2022}
Loi, M.\BBACOMMA\  \BBA\ Heitz, C. \BBOP2022\BBCP.
\newblock \BBOQ Is {Calibration} a {Fairness} {Requirement}? {An} {Argument}
  from the {Point} of {View} of {Moral} {Philosophy} and {Decision}
  {Theory}\BBCQ\
\newblock In {\Bem 2022 {ACM} {Conference} on {Fairness}, {Accountability}, and
  {Transparency}}, {FAccT} '22, \BPGS\ 2026--2034, New York, NY, USA.
  Association for Computing Machinery.

\bibitem[\protect\BCAY{Long}{Long}{2021}]{long_fairness_2021}
Long, R. \BBOP2021\BBCP.
\newblock \BBOQ Fairness in {Machine} {Learning}: {Against} {False} {Positive}
  {Rate} {Equality} as a {Measure} of {Fairness}\BBCQ\
\newblock {\Bem Journal of Moral Philosophy}, {\Bem 19\/}(1), 49--78.
\newblock Brill.

\bibitem[\protect\BCAY{Mehrabi, Morstatter, Saxena, Lerman,\ \BBA\
  Galstyan}{Mehrabi et~al.}{2022}]{mehrabi_survey_2022}
Mehrabi, N., Morstatter, F., Saxena, N., Lerman, K., \BBA\ Galstyan, A.
  \BBOP2022\BBCP.
\newblock \BBOQ A {Survey} on {Bias} and {Fairness} in {Machine}
  {Learning}\BBCQ\
\newblock {\Bem ACM Computing Surveys}, {\Bem 54\/}(6), 1--35.

\bibitem[\protect\BCAY{Meinshausen}{Meinshausen}{2006}]{meinshausen_quantile_2006}
Meinshausen, N. \BBOP2006\BBCP.
\newblock \BBOQ Quantile {Regression} {Forests}\BBCQ\
\newblock {\Bem Journal of Machine Learning Research}, {\Bem 7\/}(35),
  983--999.

\bibitem[\protect\BCAY{Mitchell, Potash, Barocas, D'Amour,\ \BBA\ Lum}{Mitchell
  et~al.}{2021}]{mitchell_prediction-based_2021}
Mitchell, S., Potash, E., Barocas, S., D'Amour, A., \BBA\ Lum, K.
  \BBOP2021\BBCP.
\newblock \BBOQ Prediction-{Based} {Decisions} and {Fairness}: {A} {Catalogue}
  of {Choices}, {Assumptions}, and {Definitions}\BBCQ\
\newblock {\Bem Annual Review of Statistics and Its Application}, {\Bem
  8\/}(1), 141--163.

\bibitem[\protect\BCAY{Nabi, Malinsky,\ \BBA\ Shpitser}{Nabi
  et~al.}{2019}]{nabi_learning_2019}
Nabi, R., Malinsky, D., \BBA\ Shpitser, I. \BBOP2019\BBCP.
\newblock \BBOQ Learning {Optimal} {Fair} {Policies}\BBCQ\
\newblock In {\Bem Proceedings of the 36th {International} {Conference} on
  {Machine} {Learning}}, \BPGS\ 4674--4682. PMLR.
\newblock ISSN: 2640-3498.

\bibitem[\protect\BCAY{Nabi\ \BBA\ Shpitser}{Nabi\ \BBA\
  Shpitser}{2018}]{nabi_fair_2018}
Nabi, R.\BBACOMMA\  \BBA\ Shpitser, I. \BBOP2018\BBCP.
\newblock \BBOQ Fair inference on outcomes\BBCQ\
\newblock In {\Bem Proceedings of the {Thirty}-{Second} {AAAI} {Conference} on
  {Artificial} {Intelligence} and {Thirtieth} {Innovative} {Applications} of
  {Artificial} {Intelligence} {Conference} and {Eighth} {AAAI} {Symposium} on
  {Educational} {Advances} in {Artificial} {Intelligence}},
  {AAAI}'18/{IAAI}'18/{EAAI}'18, \BPGS\ 1931--1940, New Orleans, Louisiana,
  USA. AAAI Press.

\bibitem[\protect\BCAY{Pan, Cui, Bian, Zhang,\ \BBA\ Wang}{Pan
  et~al.}{2021}]{pan_explaining_2021}
Pan, W., Cui, S., Bian, J., Zhang, C., \BBA\ Wang, F. \BBOP2021\BBCP.
\newblock \BBOQ Explaining {Algorithmic} {Fairness} {Through}
  {Fairness}-{Aware} {Causal} {Path} {Decomposition}\BBCQ\
\newblock In {\Bem Proceedings of the 27th {ACM} {SIGKDD} {Conference} on
  {Knowledge} {Discovery} \& {Data} {Mining}}, \BPGS\ 1287--1297, Virtual Event
  Singapore. ACM.

\bibitem[\protect\BCAY{Pearl}{Pearl}{2009}]{pearl_causality_2009}
Pearl, J. \BBOP2009\BBCP.
\newblock {\Bem Causality: {Models}, {Reasoning} and {Inference}\/} (2nd \BEd).
\newblock Cambridge University Press.

\bibitem[\protect\BCAY{Pearl, Glymour,\ \BBA\ Jewell}{Pearl
  et~al.}{2016}]{pearl_causal_2016}
Pearl, J., Glymour, M., \BBA\ Jewell, N.~P. \BBOP2016\BBCP.
\newblock {\Bem Causal {Inference} in {Statistics}}.
\newblock John Wiley \& Sons Ltd.

\bibitem[\protect\BCAY{Pedreshi, Ruggieri,\ \BBA\ Turini}{Pedreshi
  et~al.}{2008}]{pedreshi_discrimination-aware_2008}
Pedreshi, D., Ruggieri, S., \BBA\ Turini, F. \BBOP2008\BBCP.
\newblock \BBOQ Discrimination-aware {Data} {Mining}\BBCQ\
\newblock In {\Bem Proceedings of the 14th {ACM} {SIGKDD} international
  conference on {Knowledge} discovery and data mining}, {KDD} '08, \BPGS\
  560--568, New York, NY, USA. Association for Computing Machinery.

\bibitem[\protect\BCAY{Plečko, Bennett,\ \BBA\ Meinshausen}{Plečko
  et~al.}{2024}]{plecko_r_package}
Plečko, D., Bennett, N., \BBA\ Meinshausen, N. \BBOP2024\BBCP.
\newblock \BBOQ {fairadapt}: Causal reasoning for fair data preprocessing\BBCQ\
\newblock {\Bem Journal of Statistical Software}, {\Bem 110\/}(4), 1--35.

\bibitem[\protect\BCAY{Plečko\ \BBA\ Meinshausen}{Plečko\ \BBA\
  Meinshausen}{2020}]{plecko_fair_2020}
Plečko, D.\BBACOMMA\  \BBA\ Meinshausen, N. \BBOP2020\BBCP.
\newblock \BBOQ Fair {Data} {Adaptation} with {Quantile} {Preservation}\BBCQ\
\newblock {\Bem Journal of Machine Learning Research}, {\Bem 21}, 1--44.

\bibitem[\protect\BCAY{Rawls}{Rawls}{2003}]{rawls_theory_2003}
Rawls, J.~A. \BBOP2003\BBCP.
\newblock {\Bem A {Theory} of {Justice}}.
\newblock Belknap Press of Harvard Univ. Press.

\bibitem[\protect\BCAY{Roemer}{Roemer}{2018}]{roemer_theories_2018}
Roemer, J.~E. \BBOP2018\BBCP.
\newblock {\Bem Theories of {Distributive} {Justice}}.
\newblock Harvard University Press.

\bibitem[\protect\BCAY{Schwöbel\ \BBA\ Remmers}{Schwöbel\ \BBA\
  Remmers}{2022}]{schwobel_long_2022}
Schwöbel, P.\BBACOMMA\  \BBA\ Remmers, P. \BBOP2022\BBCP.
\newblock \BBOQ The {Long} {Arc} of {Fairness}: {Formalisations} and {Ethical}
  {Discourse}\BBCQ\
\newblock In {\Bem 2022 {ACM} {Conference} on {Fairness}, {Accountability}, and
  {Transparency}}, \BPGS\ 2179--2188, Seoul Republic of Korea. ACM.

\bibitem[\protect\BCAY{Selbst, Boyd, Friedler, Venkatasubramanian,\ \BBA\
  Vertesi}{Selbst et~al.}{2019}]{selbst_fairness_2019}
Selbst, A.~D., Boyd, D., Friedler, S.~A., Venkatasubramanian, S., \BBA\
  Vertesi, J. \BBOP2019\BBCP.
\newblock \BBOQ Fairness and {Abstraction} in {Sociotechnical} {Systems}\BBCQ\
\newblock In {\Bem Proceedings of the {Conference} on {Fairness},
  {Accountability}, and {Transparency}}, \BPGS\ 59--68, Atlanta GA USA. ACM.

\bibitem[\protect\BCAY{Singh, Kempe,\ \BBA\ Joachims}{Singh
  et~al.}{2021}]{singh_fairness_2021}
Singh, A., Kempe, D., \BBA\ Joachims, T. \BBOP2021\BBCP.
\newblock \BBOQ Fairness in {Ranking} under {Uncertainty}\BBCQ\
\newblock In Ranzato, M., Beygelzimer, A., Dauphin, Y., Liang, P.~S., \BBA\
  Vaughan, J.~W.\BEDS, {\Bem Advances in {Neural} {Information} {Processing}
  {Systems}}, \lowercase{\BVOL}~34, \BPGS\ 11896--11908. Curran Associates,
  Inc.

\bibitem[\protect\BCAY{Suresh\ \BBA\ Guttag}{Suresh\ \BBA\
  Guttag}{2021}]{suresh_framework_2021}
Suresh, H.\BBACOMMA\  \BBA\ Guttag, J.~V. \BBOP2021\BBCP.
\newblock \BBOQ A {Framework} for {Understanding} {Sources} of {Harm}
  throughout the {Machine} {Learning} {Life} {Cycle}\BBCQ\
\newblock \textit{Equity and Access in Algorithms, Mechanisms, and
  Optimization}, 1--9.

\bibitem[\protect\BCAY{Thorndike}{Thorndike}{1971}]{thorndike_concepts_1971}
Thorndike, R.~L. \BBOP1971\BBCP.
\newblock \BBOQ Concepts of {Culture}-{Fairness}\BBCQ\
\newblock {\Bem Journal of Educational Measurement}, {\Bem 8\/}(2), 63--70.

\bibitem[\protect\BCAY{Verma\ \BBA\ Rubin}{Verma\ \BBA\
  Rubin}{2018}]{verma_fairness_2018}
Verma, S.\BBACOMMA\  \BBA\ Rubin, J. \BBOP2018\BBCP.
\newblock \BBOQ Fairness {Definitions} {Explained}\BBCQ\
\newblock In {\Bem Proceedings of the {International} {Workshop} on {Software}
  {Fairness}}, Gothenburg Sweden. ACM.

\bibitem[\protect\BCAY{Wachter}{Wachter}{2019}]{wachter_affinity_2019}
Wachter, S. \BBOP2019\BBCP.
\newblock \BBOQ Affinity {Profiling} and {Discrimination} by {Association} in
  {Online} {Behavioural} {Advertising}\BBCQ\
\newblock {\Bem Berkeley Technology Law Journal}, {\Bem 35\/}(2).

\bibitem[\protect\BCAY{Wachter, Mittelstadt,\ \BBA\ Russell}{Wachter
  et~al.}{2021}]{wachter_bias_2021}
Wachter, S., Mittelstadt, B., \BBA\ Russell, C. \BBOP2021\BBCP.
\newblock \BBOQ Bias {Preservation} in {Machine} {Learning}: {The} {Legality}
  of {Fairness} {Metrics} {Under} {EU} {Non}-{Discrimination} {Law}\BBCQ\
\newblock {\Bem West Virginia Law Review}, {\Bem 123\/}(3), 735--790.

\bibitem[\protect\BCAY{Wong}{Wong}{2020}]{wong_democratizing_2020}
Wong, P.-H. \BBOP2020\BBCP.
\newblock \BBOQ Democratizing {Algorithmic} {Fairness}\BBCQ\
\newblock {\Bem Philosophy \& Technology}, {\Bem 33\/}(2), 225--244.

\bibitem[\protect\BCAY{Zwick\ \BBA\ Dorans}{Zwick\ \BBA\
  Dorans}{2016}]{zwick_philosophical_2016}
Zwick, R.\BBACOMMA\  \BBA\ Dorans, N.~J. \BBOP2016\BBCP.
\newblock \BBOQ Philosophical {Perspectives} on {Fairness} in {Educational}
  {Assessment}\BBCQ\
\newblock In {\Bem Fairness in educational assessment and measurement}, \BPGS\
  283--298. Routledge.

\end{thebibliography}
\bibliographystyle{theapa}

\end{document}